\documentclass{article}

\usepackage{amsmath,amssymb,amsfonts}
\usepackage[square,numbers,sort&compress]{natbib}
\usepackage{authblk}
\usepackage{algorithmic}
\usepackage{graphicx}
\usepackage{textcomp}
\usepackage{hyperref}
\hypersetup{
    hidelinks
}
\usepackage{tabularx}
\usepackage{caption}
\usepackage{subcaption}
\usepackage{xcolor}

\providecommand{\keywords}[1]
{
  \small	
  \textbf{\textit{Keywords---}} #1
}

\title{TEMPL: A Resource-Efficient and Transparent Model-Free Framework for Pseudo-Label Generation in Material CT Scans}

\author[1,*]{Ashiqur Rahman}
\author[1]{Venkata Devesh Reddy Seethi}
\author[2]{Austin Yunker}
\author[2]{Rajkumar Kettimuthu}
\author[1]{Hamed Alhoori}

\affil[1]{Department of Computer Science, Northern Illinois University, DeKalb, IL, USA}
\affil[2]{Argonne National Laboratory, Lemont, IL, USA}
\affil[*]{Corresponding author: ashiqur.r@niu.edu}

\date{}

\begin{document}

\maketitle

%% Abstract
\begin{abstract}
The scarcity of large annotated datasets is a primary obstacle to deploying deep learning (DL) models for segmentation in industrial material CT analysis. Current DL-based pseudo-labeling techniques attempt to mitigate this issue, but often remain computationally expensive and lack interpretability. This paper introduces a multi-stage model-free framework designed to generate high-quality pseudo-labels directly from grayscale CT scans. By integrating adaptive thresholding with a custom breadth-first search region-growing algorithm, our approach produces reliable polygonal annotations without requiring pre-trained weights or GPU acceleration. We evaluated this framework against the zero-shot Segment Anything Model (SAM), a fully supervised 3D U-Net, and classical baselines. Our analysis reveals that while foundation models may highlight relevant pixels, they struggle to maintain instance-level structural integrity by hallucinating noise and merging adjacent particles. In quantitative validation against ground truth, this translated to an IoU of only $10.74\%$ for the zero-shot foundation model, compared to $63.19\%$ achieved by our model-free approach despite operating with no labeled training data. Conversely, our model-free pipeline captures a physically accurate representation of the material features. Downstream validation confirms this utility, showing that our generated labels successfully facilitate end-to-end model training, while our microstructural analysis provides rigorous quantitative validation of the extracted physical parameters. The microstructural quantification experiment shows that grains extracted by our method exhibit a median equivalent diameter of $163.8 {\mu}m$, closely tracking the specimen's known $106-150 {\mu}m$ sieve range, compared to a coarser $219.5 {\mu}m$ median produced by SAM. Furthermore, our method operates entirely on a standard CPU in nearly half the processing time of GPU-accelerated alternatives. These findings establish our transparent approach as a highly effective tool for annotation bootstrapping and microstructural quantification in resource-limited environments.
\end{abstract}


% Keywords
\keywords{Computed tomography, Microstructure analysis, Material characterization, Pseudo labeling, Image segmentation, Resource efficient computing}

\section{Introduction} \label{sec:pseudo-introduction}
The scarcity of large, accurately labeled datasets presents one of the most significant obstacles to deploying robust object detection and image segmentation models in diverse industrial environments. This bottleneck is especially pronounced in integrated computational materials engineering (ICME) workflows, where microstructural quantification depends on pixel-accurate segmentation of CT volumes. As one concrete example, producing the expert-annotated ground truth for the dendritic microstructure dataset used later in this study has been reported to require approximately 20 minutes per image slice \citep{James2023-dc}, a cost that scales prohibitively across the thousands of slices contained in a typical volumetric CT scan. The same constraint extends well beyond materials science, where manual annotation by domain experts is more generally recognized as a costly bottleneck \citep{Candemir2021-nu}. Data variability further complicates model generalization across different settings \citep{Liang2020-zi, Zhang2020-jj}. To mitigate these challenges, the research community has increasingly gravitated towards semi-supervised and unsupervised learning, dependent on pseudo-labeling to leverage unlabeled datasets and reduce reliance on expert labor \citep{Kage2024-cg, Niu2024-qj, Wang2024-mu, Ran2025-jr, Rickmann2025-jf}. However, the success of these methods depends on label quality, which remains vulnerable to noise and confirmation bias \citep{Seibold2022-vi, Su2024-ud}.

State-of-the-art pseudo-labeling methods often use sophisticated deep learning (DL) models, including large foundation models, to mitigate confirmation bias \citep{Qiu2024-vp, Su2024-ud, Ran2025-jr, Huang2023-wr, Kim2024-yq, Yang2024-bo, Rickmann2025-jf}. However, these DL approaches are typically computationally expensive and lack explainability. As an alternative, traditional model-free techniques like thresholding \citep{Otsu1979-jv}, edge detection \citep{Rana2014-mq}, and clustering \citep{Bhattacharjee2021-zf} offer a less resource-intensive path. Applied in isolation, these methods are often fragile, and their sensitivity to image variations limits their generalizability \citep{Huang2024-ib, Saleh2025-kf}. No widely adopted framework currently exists that combines these classical techniques into a robust, interpretable pipeline specifically designed for the challenging intensity profiles and complex morphologies of grayscale material CT images.

% 
% This paper addresses that gap by introducing a multi-stage model-free framework for pseudo-label generation tailored for material CT scans.

This paper addresses that gap by proposing \textbf{TEMPL}, a \textbf{T}ransparent and \textbf{E}fficient \textbf{M}odel-free \textbf{P}seudo-\textbf{L}abeling framework, which introduces a multi-stage model-free approach to pseudo-label generation tailored for material CT scans. Rather than relying on any single heuristic, our approach integrates adaptive thresholding, morphological operations, and a custom region-growing algorithm into a coherent pipeline that identifies objects and converts them into polygonal annotations. The core of the framework is a custom region-growing algorithm, which extends the seeded region growing paradigm, employs an automated local seed identification requiring no prior phase specification, and BFS-based propagation specifically designed to handle discontinuous boundaries and heterogeneous intensity profiles characteristic of material CT data. This composition overcomes the fragility of individual classical methods, reducing sensitivity to pixel intensity variations while preserving the computational efficiency and interpretability that DL-based methods sacrifice. These properties make our approach particularly well-suited for grayscale material CT images, where data scarcity and complex object morphologies limit the practical applicability of deep learning.

We validate our method through a mixed-method evaluation involving quantitative and qualitative assessments, benchmarked against the zero-shot Segment Anything Model (SAM) \citep{Kirillov2023-dh}, a fully supervised 3D U-Net \citep{James2023-dc}, and classical baseline methods. The quantitative evaluation reveals an important distinction in how different approaches perform as pseudo-label generators. While a zero-shot foundation model may highlight relevant pixels, our experiments reveal that it struggles to maintain instance-level structural integrity. It simultaneously hallucinates instances from background noise and merges distinct, adjacent particles into artificially large structures, making it insufficient as a standalone annotation tool for data-scarce material imaging. In contrast, our framework produces far more reliable and physically accurate signals suitable for downstream model training. Unlike fully supervised approaches, which require substantial quantities of manually annotated training data that are costly and time-intensive to produce in specialized domains \citep{James2023-dc}, the proposed framework generates pseudo-labels without any labeled examples, operating directly on raw grayscale CT images. The qualitative analysis confirms this practical utility, showing that models trained on our pseudo-labels achieve object detection and instance segmentation performance comparable to those trained on SAM-generated labels, while our method runs nearly twice as fast on a standard CPU without requiring GPU resources. Additionally, a comparative experiment on microstructural quantification proves the practical utility of our approach in capturing accurate physical parameters for materials science. These findings position our framework as a fast, transparent, and resource-efficient tool for annotation bootstrapping in scientific and industrial environments where labeled data and computing infrastructure remain limited.

\section{Related Works} \label{sec:pseudo-related-works}
\subsection{Image Segmentation and Object Detection}\label{sec:pseudo-image-segmentation-detection}

Image segmentation, a cornerstone of computer vision, involves partitioning a digital image into multiple, semantically meaningful segments. From traditional pixel classification techniques, image segmentation evolved into more advanced DL-based approaches with high object localization accuracy and instance differentiation ability \citep{Hafiz2020-id}. Modern image segmentation is dominated by supervised deep learning models, which have demonstrated state-of-the-art performance across numerous domains. Architectures based on Convolutional Neural Networks (CNNs), particularly Fully Convolutional Networks (FCNs), set the stage by enabling end-to-end, pixel-wise semantic segmentation \citep{Long2015-cp}. The U-Net architecture, with its characteristic encoder-decoder structure and skip connections, became a standard in medical and materials imaging \citep{Ronneberger2015-yp, Siddique2021-ra}. With further refinements from innovations like U-Net++ improved its ability to handle objects of varying scales and shapes \citep{Zhou2018-vo}.

These advancements have widespread applications in domains like medical imaging for computer-aided diagnosis \citep{Su2011-re, Irfan2021-ra, Geng2022-ax} and industrial defect detection \citep{Sreeshan2020-dc, Rahman2025-wf, Daoud2019-va, Wang2025-ip}. Landmark models have continuously pushed state-of-the-art performance, including Mask R-CNN for instance segmentation \citep{He2017-we}, the YOLO family for real-time object detection \citep{Redmon2016-ap, Khanam2024-hp}, and the promptable, zero-shot Segment Anything Model (SAM) \citep{Kirillov2023-dh}.

Despite these successes, the reliance on high-quality labeled data remains a significant bottleneck, particularly in specialized domains like materials science and medical imaging, where manual annotation is laborious, expensive, and requires deep domain expertise \citep{Cai2017-gv, He2017-we, Daoud2019-va, Hafiz2020-id, Candemir2021-nu}.

\subsection{Unsupervised Models}\label{sec:pseudo-unsupervised-models}
To address the data-labeling bottleneck, researchers have turned to unsupervised and semi-supervised learning to leverage the vast amounts of unlabeled data. However, achieving high accuracy and generalizability without prohibitive computational cost remains a central challenge \citep{Su2011-re, Xia2017-rd, Wang2023-dp, Chun2024-vj, Liu2024-kk, Wang2024-mu}.

Several fully unsupervised models have been proposed to overcome the limitations of classical methods, which struggle with noise and texture \citep{Zhang2008-gm}. DL-based unsupervised approaches include W-Net \citep{Xia2017-rd}, a fully convolutional autoencoder, and CUTS\citep {Liu2024-kk}, which performs comparably to foundation models like SAM \citep{Kirillov2023-dh, Mazurowski2023-ts} and its medical variants \citep{Ye2023-ms, Ma2024-od} without requiring large labeled datasets. Other recent frameworks like U2Seg \citep{Niu2024-qj} and UnSAM \citep{Wang2024-mu} internally generate pseudo-labels via clustering or decomposition to achieve competitive performance on universal segmentation tasks.

A common thread in these advanced unsupervised models is their reliance on pseudo-label generation within their training pipeline. The complexity and generalizability of this step across different domains remain open questions \citep{Liang2020-zi, Zhang2020-jj, Bai2025-yg}, making the development of efficient pseudo-labeling methods a critical research avenue for improving the practicality of unsupervised models.

\subsection{Pseudo-label Generation}\label{sec:pseudo-pseudo-label-generation}
Pseudo-labeling methodologies in image segmentation generally bifurcate into model-free and deep learning-based (DL) approaches. Model-free techniques utilize pre-defined algorithms grounded in image properties like intensity or morphology. While these methods, such as K-Means clustering \citep{Liu2022-fk}, the Watershed algorithm \citep{Kornilov2022-hj}, or simple heuristics \citep{Yang2022-kf}, are computationally efficient and interpretable, these approaches remain vulnerable to image noise and stochastic variations \citep{Chen2021-cj, Militello2021-la}. Region-growing methods have been applied to materials CT segmentation of granular media, including multi-phase geomaterial imaging \citep{Cheng2020-sb}, where seed regions and intensity thresholds are defined per known material phase. The foundational seeded region-growing formulations \citep{Adams1994-hz, Mehnert1997-wr, Hojjatoleslami1998-md} rely on manually selected seeds and conventional connectivity assumptions, and their application to materials CT has retained the requirement for prior phase specification. These constraints limit applicability to unlabeled, heterogeneous monochromatic CT data where phase boundaries are discontinuous and intensity profiles vary within individual particles.

Beyond thresholding, several other classical model-free techniques are widely used for materials and porous-media segmentation, though each carries limitations that constrain standalone use on grayscale material CT data. Watershed-based segmentation is highly sensitive to local intensity noise, where individual pixel-level minima are misinterpreted as distinct catchment basins, producing a large number of spurious regions unless mitigated through marker-controlled pre-processing or post-hoc merging \citep{Kornilov2022-hj, Schluter2014-jb}. Superpixel-based approaches such as SLIC \citep{Achanta2012-tf} improve computational efficiency by grouping pixels into compact, perceptually coherent clusters, but their reliance on local color or intensity similarity limits performance in low-contrast regions where object boundaries are poorly defined, a characteristic common to grayscale material CT data \citep{Wildenschild2013-nn}. Active contour and level set methods offer greater flexibility for complex, non-convex object shapes, but this flexibility comes at a cost. Segmentation quality depends strongly on the placement of the initial contour, evolution is prone to converging on a local minimum rather than the true object boundary, and the iterative curve evolution required at each step is computationally expensive relative to single-pass thresholding or region-growing approaches \citep{Chen2021-rk}. Even recent models targeting industrial CT volumes directly continue to require carefully engineered formulations to manage noise, artifacts, and intensity inhomogeneity in complex scanned components \citep{Li2026-bv}, underscoring that these challenges are not solved by minor parameter tuning within the active contour framework itself. Applied individually, no single classical technique reliably handles the combination of discontinuous boundaries, heterogeneous intensity profiles, and noise present in material CT data, motivating the composite pipeline design introduced in this work.

Conversely, modern DL-based approaches are better suited for capturing complex features. A prevalent strategy is iterative self-training, where models refine performance by retraining on their own predictions \citep{Rickmann2025-jf}. To enhance label fidelity, researchers have introduced strategies such as temporal ensembling, clustering consensus, and confidence thresholding \citep{Zhang2021-xm, Wang2021-ml}. However, single-model architectures are often prone to confirmation bias, reinforcing initial prediction errors \citep{Seibold2022-vi, Su2024-ud}. To counteract this, advanced frameworks employing teacher-student architectures or mutual training have been developed \citep{Seibold2022-vi, Qiu2024-vp, Su2024-ud, Toba2024-bh, Ran2025-jr}.

Recently, large foundation models like SAM have emerged as potent zero-shot pseudo-label generators, often achieving performance comparable to fully supervised systems \citep{Huang2023-wr, Kim2024-yq, Yang2024-bo, Rickmann2025-jf}. Despite their efficacy, their substantial computational overhead remains a significant barrier for many applications. Building upon the inherent efficiency and interpretability of model-free generation, this work proposes a cost-effective algorithm that integrates several heuristic approaches. Our method provides a rapid, transparent benchmark for the development and evaluation of image segmentation models.

\section{Dataset} \label{sec:pseudo-dataset}
This study utilizes two distinct datasets. The primary dataset consists of unlabeled sand grain CT images used for initial processing and pseudo-label generation. We also employ a secondary dataset to quantitatively validate the algorithm against established ground-truth annotations.

\begin{figure}[!hbt]
  \begin{subfigure}[b]{0.48\linewidth}
    \includegraphics[width=\linewidth]{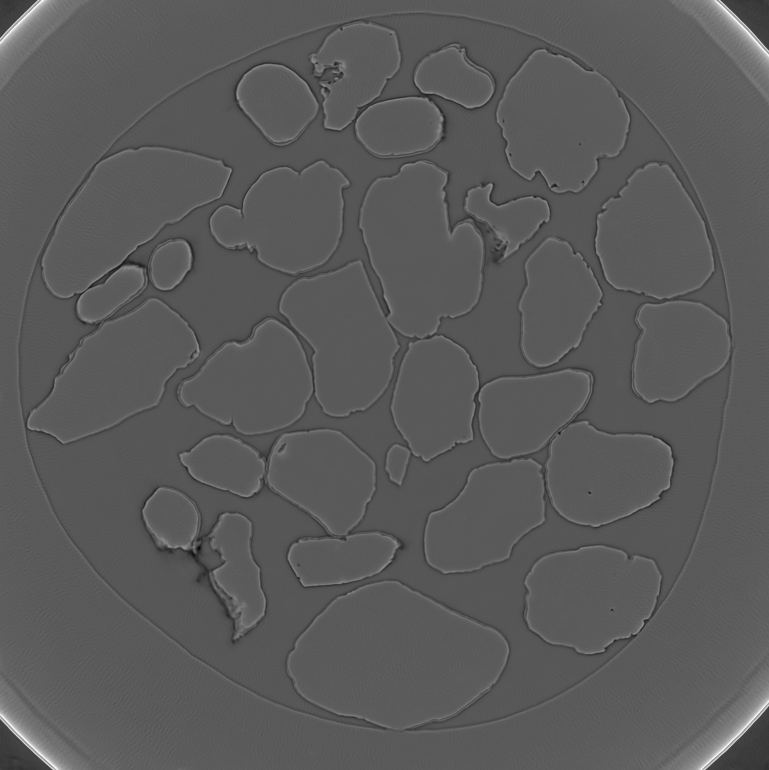}
    \caption{A sample image of the scanned sand grains.}
    \label{fig:pseudo-sample-sand-grain}
  \end{subfigure}
  \hfill
  \begin{subfigure}[b]{0.48\linewidth}
    \includegraphics[width=\linewidth]{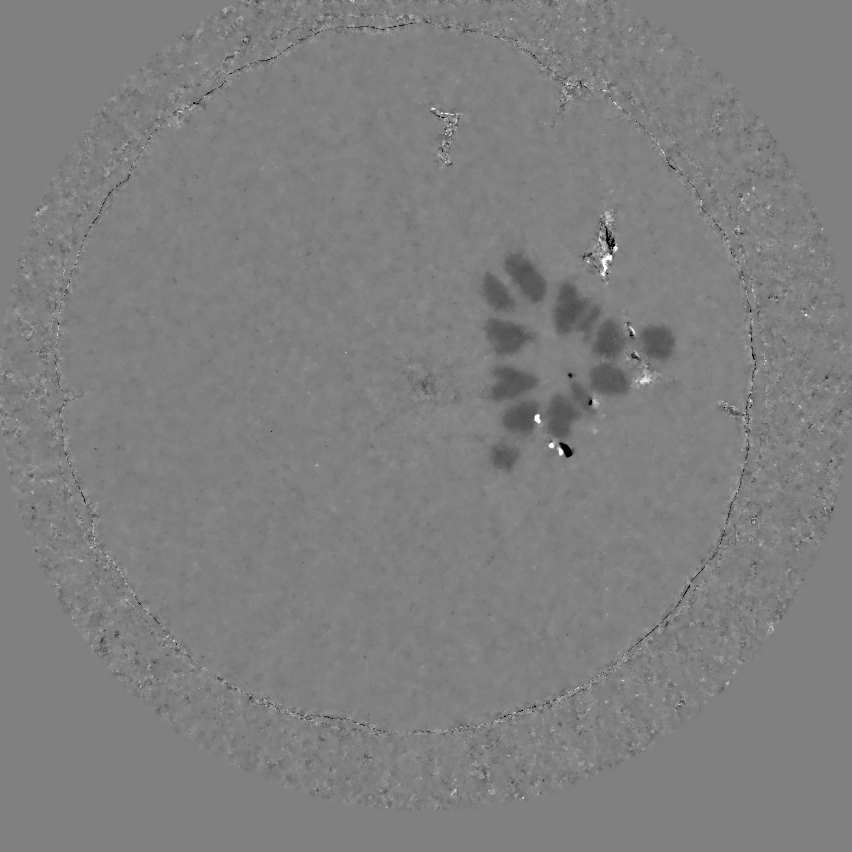}
    \caption{A sample image from the Dendrites dataset.}
    \label{fig:pseudo-sample-dendrite}
  \end{subfigure}
  \caption{Representative samples from the datasets.}
\end{figure}

\subsection{Primary Dataset} \label{sec:pseudo-primary-dataset}
The primary data consists of a silica sand dataset featuring $2705$ particles analyzed under mechanical compression \citep{Imseeh2020-az}. These data were acquired using a monochromatic $72$ keV X-ray beam with dimensions of $1.6 \times 1.2$ mm and a $360^\circ$ rotation at $0.2^\circ$ increments. The resulting three-dimensional volume comprises $1200$ slices, each with a resolution of $1520 \times 1520$ pixels, which were further denoised.

We further refined the dataset for our study. We excluded slices lacking discernible sand grain features and resized the remaining images to enhance computational efficiency. The final dataset consists of $1192$ grayscale images at $769 \times 770$ resolution, capturing grains with diverse intensity profiles. A representative sample is illustrated in Fig. \ref{fig:pseudo-sample-sand-grain}.

\subsection{Validation Dataset} \label{sec:pseudo-validation-dataset}
Quantitative validation was performed using the Dendrites dataset, which was acquired via synchrotron-based $4D$ X-ray Computed Tomography at Argonne National Laboratory \citep{Stan2019-kg, Stan2020-ys}. The volumetric data were partitioned into $2D$ z-slices to facilitate the segmentation of solid and liquid phases. This dataset provides the ground-truth labels required for an objective evaluation of our model-free outputs. Fig. \ref{fig:pseudo-sample-dendrite} displays a representative sample from this validation set.

\section{Method} \label{sec:pseudo-method}
\begin{figure}[!htb]
    \centering
    \includegraphics[width=\linewidth]{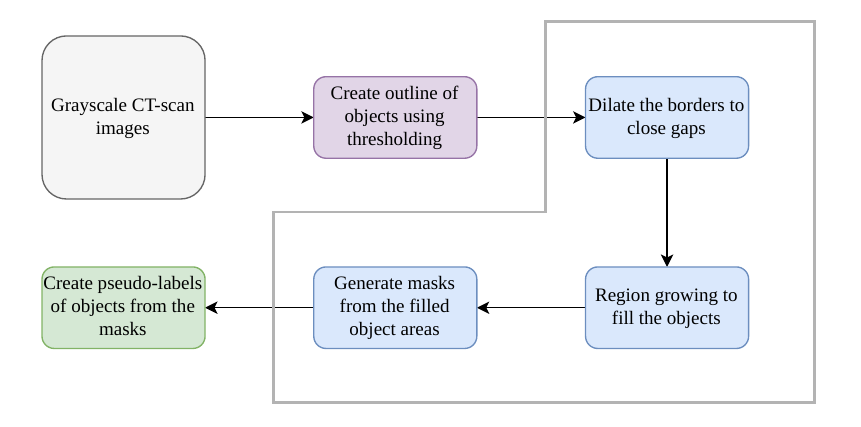}
    \caption{Steps to create pseudo label from CT-scan images}
    \label{fig:pseudo-pseudo-label-creation-steps}
\end{figure}

Our pseudo-label generation technique is a multi-stage pipeline, as illustrated in Fig. \ref{fig:pseudo-pseudo-label-creation-steps}. The process consists of two primary phases. In the first phase, highlighted in purple, we use classical image processing to generate preliminary outlines of the sand grain regions from the raw grayscale images. The second phase, enclosed in the gray bounding box, involves enhancing these outlines to create the final annotations through a sequence of morphological dilation, a custom region-growing algorithm to create solid masks, and finally, the conversion of these masks into polygonal pseudo-labels.

\begin{figure}[!hbt]
  \begin{subfigure}[b]{0.32\linewidth}
    \includegraphics[width=\linewidth]{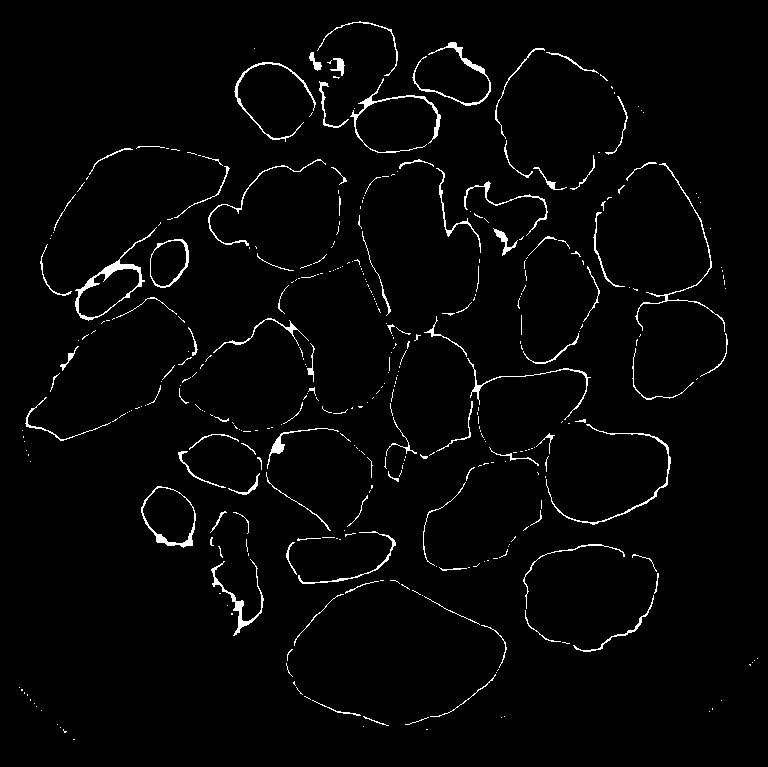}
    \caption{Outlines from the threshold.}
    \label{fig:pseudo-cleared-outline}
  \end{subfigure}
  \hfill
  \begin{subfigure}[b]{0.32\linewidth}
    \includegraphics[width=\linewidth]{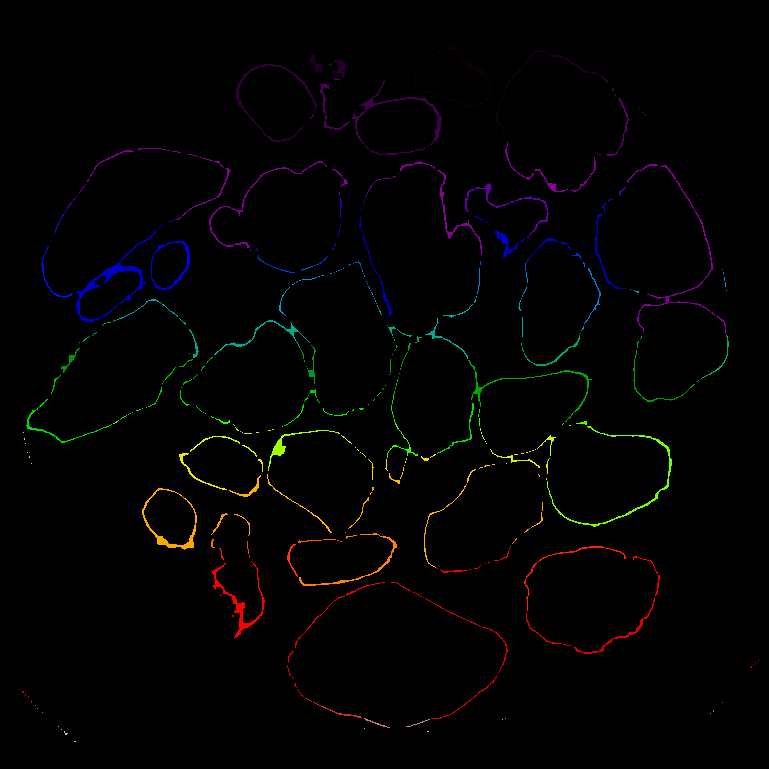}
    \caption{Labels from the outlines.}
    \label{fig:pseudo-generated-labels}
  \end{subfigure}
  \hfill
  \begin{subfigure}[b]{0.32\linewidth}
    \includegraphics[width=\linewidth]{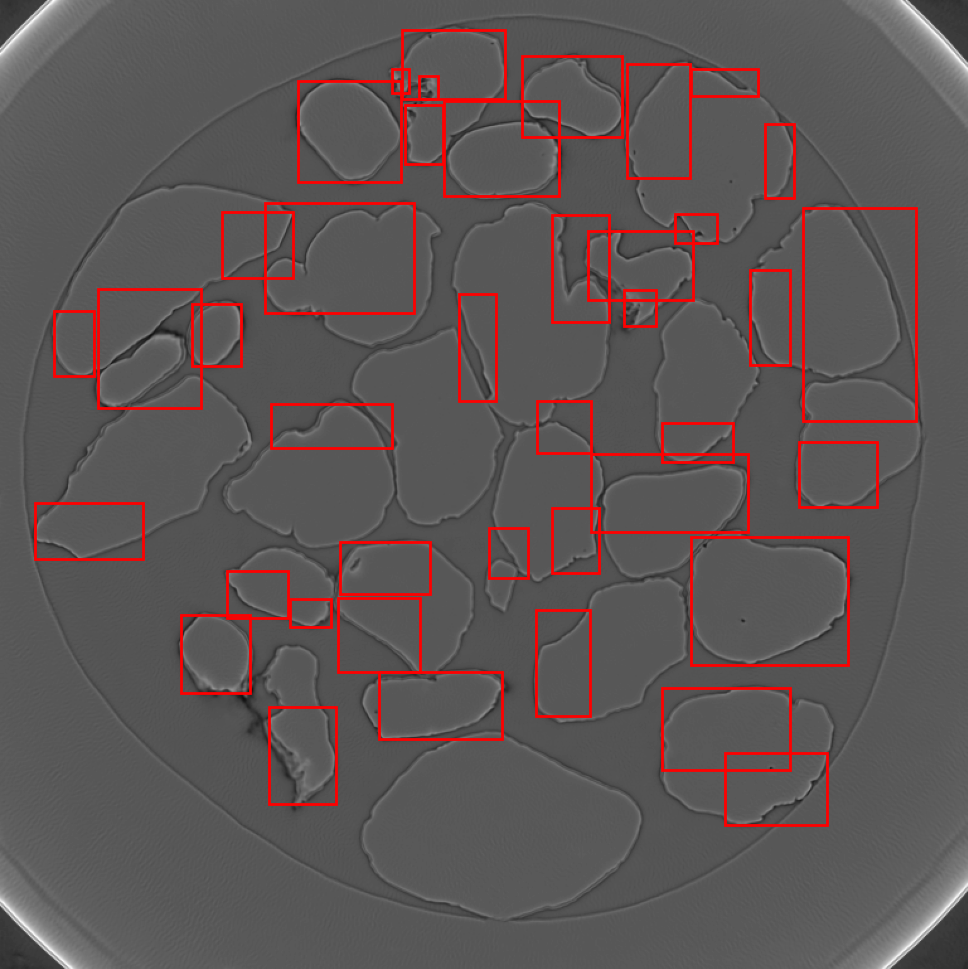}
    \caption{Regions of different labels.}
    \label{fig:pseudo-generated-regions}
  \end{subfigure}
  \caption{Threshold-based label mask generated from the original image.}
\end{figure}

\subsection{Image Processing} \label{sec:pseudo-image-processing}
The initial stage of our method creates a preliminary segmentation of the sand grains. We first apply Otsu's method \citep{Otsu1979-jv} to the grayscale images to create a binary outline of the objects, as shown in Fig. \ref{fig:pseudo-cleared-outline}. We select Otsu's method for this initial binarization because it computes a threshold automatically by maximizing the variance between foreground and background pixel intensities, requiring no manual calibration or labeled training data. This property suits material CT imaging, where intensity characteristics can shift from slice to slice due to acquisition effects such as beam hardening, making a single manually tuned threshold impractical across a large volumetric dataset. Following binarization, we use connected-component labeling\footnote{https://scikit-image.org/docs/0.25.x/api/skimage.segmentation.html} to assign a unique label to each contiguous foreground region, as seen in Fig. \ref{fig:pseudo-generated-labels}. To reduce noise, any labeled region smaller than a \(10 \times 10\) pixel area is discarded. At the dataset's spatial resolution of $1.17 {\mu}m/pixel$, this corresponds to roughly $11.7 \times 11.7 {\mu}m$, well below the smallest expected physical grain size of $106 {\mu}m$ in this specimen. The filter therefore removes isolated noise pixels and thresholding artifacts rather than legitimate grains, and the value can be adjusted directly for datasets with a different particle size distribution or spatial resolution. The remaining significant regions are displayed in Fig. \ref{fig:pseudo-generated-regions}. Because Otsu thresholding applies a single global intensity boundary, it cannot reliably resolve object boundaries where local contrast is low, such as at grain to grain contact points or regions affected by partial volume effects. This initial process therefore results in fragmented labels for what should be single sand grains, a challenge addressed directly by the dilation and region growing steps described in Section \ref{sec:pseudo-label-generation}.

\subsection{Label Generation} \label{sec:pseudo-label-generation}
The next stage of our algorithm transforms the fragmented label maps from the initial processing into complete, high-quality pseudo-labels. The refinement is achieved through a four-step sequence that includes dilation to close small boundary gaps, label creation to merge the now-contiguous regions, region growing to fill each object's interior, and refinement to clean the resulting masks and convert them into polygonal annotations. Each step targets a specific limitation of the preceding stage and exposes a small number of tunable parameters, so that any resulting error can be traced to its source rather than treated as an opaque failure of the overall pipeline.

\subsubsection*{Step 1 - Dilation}
A key challenge of the initial thresholding process is the presence of discontinuities and gaps in the boundaries of the segmented objects. To rectify this, we apply a morphological dilation\footnote{https://scikit-image.org/docs/0.25.x/api/skimage.morphology.html} operation to the binary outlines, using a $3 \times 3$ rectangular structuring element applied uniformly across the image. This expands the foreground regions by a single pixel in each direction, effectively closing minor gaps and ensuring that each object has a continuous, solid border. The kernel size represents an explicit trade-off between gap closure and region separation. A larger structuring element bridges wider discontinuities but increases the risk of merging genuinely adjacent particles into a single region, while a smaller element preserves separation between close objects but may leave larger gaps unresolved. We use a minimal $3 \times 3$ kernel because it closed the majority of observed boundary discontinuities in this dataset without producing visible merges between neighboring grains. This kernel size can be adjusted directly to accommodate material samples with different characteristic gap sizes or particle spacing. The result of this step is illustrated in Fig. \ref{fig:pseudo-dilated-outline}.

\subsubsection*{Step 2 - Create Labels}
\begin{figure}[!hbt]
  \begin{subfigure}[b]{0.32\linewidth}
    \includegraphics[width=\linewidth]{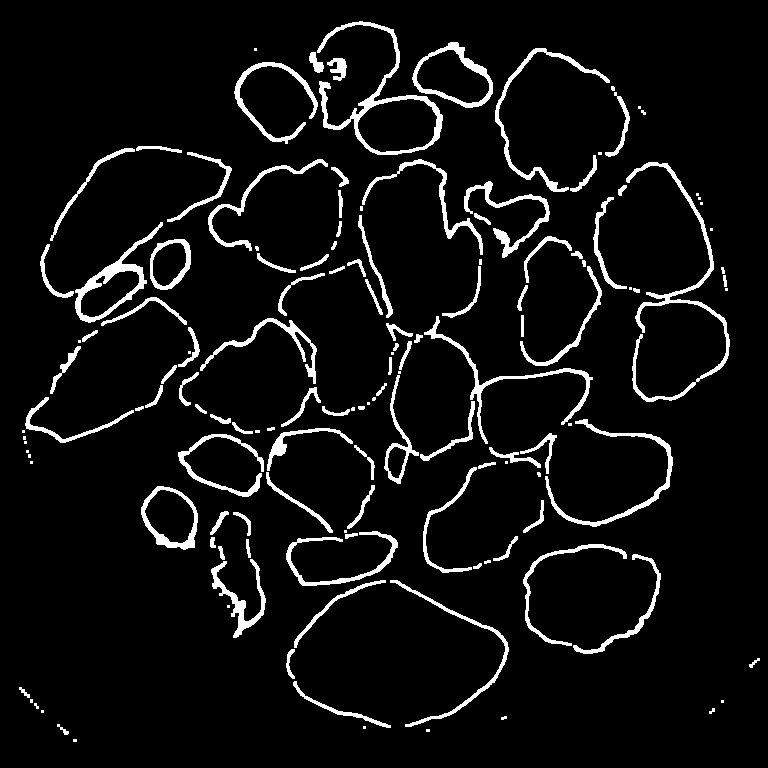}
    \caption{Outlines generated after dilation.}
    \label{fig:pseudo-dilated-outline}
  \end{subfigure}
  \hfill
  \begin{subfigure}[b]{0.32\linewidth}
    \includegraphics[width=\linewidth]{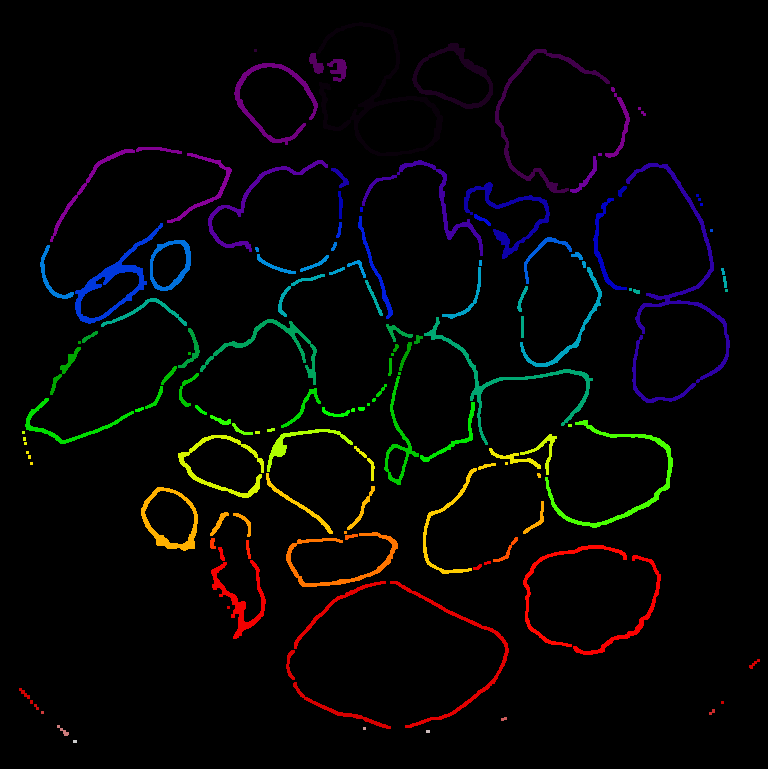}
    \caption{Labels from the dilated outlines.}
    \label{fig:pseudo-dilated-labels}
  \end{subfigure}
  \hfill
  \begin{subfigure}[b]{0.32\linewidth}
    \includegraphics[width=\linewidth]{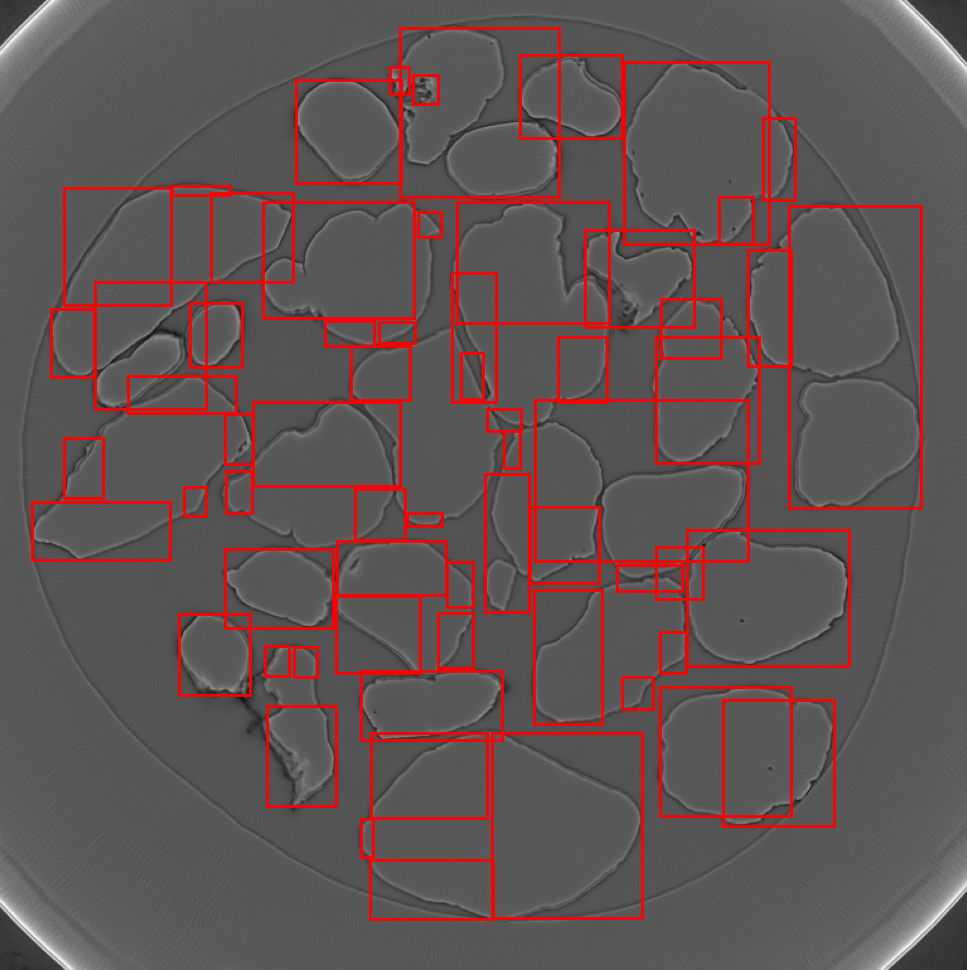}
    \caption{Regions of labels from dilated outlines.}
    \label{fig:pseudo-dilated-regions}
  \end{subfigure}
  \caption{Label mask generated from the dilated outlines.}
\end{figure}

With the dilated object boundaries from the previous step, we reapply connected-component labeling to the image, using full connectivity\footnote{https://scikit-image.org/docs/0.25.x/api/skimage.measure.html} to define adjacent pixels, meaning two pixels are treated as part of the same region if they share either an edge or a corner. We select full connectivity here because the minimal $3 \times 3$ dilation from Step $1$ sometimes closes a gap so that two fragments touch only at a single corner pixel. Full connectivity treats these corner-touching pixels as one continuous object, while edge-only connectivity would still read them as two separate ones. This step is critical as it merges many of the previously fragmented regions that now form contiguous areas, which are visually evident when comparing the resulting label map, shown in Fig. \ref{fig:pseudo-dilated-labels}, and the regions, shown in Fig. \ref{fig:pseudo-dilated-regions}, with the Figures Fig. \ref{fig:pseudo-generated-labels} and Fig. \ref{fig:pseudo-generated-regions} from previous stage.

\subsubsection*{Step 3 - Region Growing}
The next step is to create a solid mask for each object by filling its interior. To accomplish this, we developed a custom region-growing algorithm that operates in two main phases, seed identification and mask generation using Breadth-First Search (BFS).

\textbf{Seed Identification.} The process begins with a raster scan of the image. When the scan encounters a pixel \((r, c)\) that is part of a dilated boundary mask from the previous step, it initiates a search to find a representative ``seed" pixel from the object's interior. The goal is to find the characteristic pixel value of the region enclosed by the boundary.

Let the image have dimensions \(H \times W\), height and width respectively, and the value of a pixel at \((i, j)\) be \(P_{i,j}\). The algorithm first defines a search space \(S\) consisting of all pixels that appear after the current boundary pixel \((r, c)\) in the scan order. This process is defined by the Equation \ref{eq:pseudo-calculating-s}.

% \begin{multline}
\begin{equation}
\label{eq:pseudo-calculating-s}
    S = \{(i,j) | (0 \leq i < H) \wedge (0 \leq j < W) \wedge ((i > r) \vee (i = r \wedge j > c))\}
\end{equation}
% \end{multline}

Next, it identifies the subset of these pixels, \(S_{diff}\), whose values differ from that of the current boundary pixel, \(P_{current}\). This ensures the search for a seed pixel remains confined within the current region of interest, preventing it from extending beyond the object's far boundary.

\begin{equation}
    \label{eq:pseudo-calculating-s-diff}
    S_{diff} = \{(i,j) \in S | P_{i,j} \neq P_{current}\}
\end{equation}

From this subset \(S_{diff}\), the algorithm selects the very first pixel it encounters, \((r^*, c^*)\), and treats it as the definitive interior seed for the object. The intensity value of this seed pixel, \(P_{r^*,c^*}\), is designated as the reference value \(P_{ref}\) for the entire region. In the unlikely event that \(S_{diff}\) is empty, meaning the rest of the image has the same intensity as the boundary, the algorithm defaults to using the boundary's own value, \(P_{current}\), as the reference. This process is formalized in Equation \ref{eq:pseudo-calculating-p-new}.

\begin{equation}
    \label{eq:pseudo-calculating-p-new}
    P_{ref} = \begin{cases}
        P_{r^*,c^*} & if S_{diff} \neq \phi \wedge (r^*, c^*) \in S_{diff} \\
        P_{current} & if S_{diff} = \phi
    \end{cases}
\end{equation}

\textbf{Mask Generation.} Once the reference value \(P_{ref}\) is established, a Breadth-First Search (BFS) is initiated from the seed pixel's location to identify and flag all connected pixels that belong to the object's interior. The search explores a \(5 \times 5\) pixel neighborhood around each pixel in the queue, excluding the center pixel itself. The neighborhood, defined by the offsets in \(\Delta^*\) in Equation \ref{eq:pseudo-delta-set}, is deliberately limited to a \(2-pixel\) reach in any direction to prevent the search from accidentally leaking into adjacent but separate object regions.

\begin{equation}
\label{eq:pseudo-delta-set}
    \Delta^* = \{-2, -1, 0, 1, 2\}^2 \setminus {(0,0)}
\end{equation}

For a given pixel \((r, c)\), the set of neighbors N to be evaluated is formalized by the Equation \ref{eq:pseudo-neighbor-set}.

\begin{equation}
\label{eq:pseudo-neighbor-set}
    N = {(r + dr, c + dc) | (dr, dc) \in \Delta^*}
\end{equation}

A neighboring pixel at \((nr, nc)\) is classified as part of the object mask if it satisfies a set of conditions, \(C(nr, nc)\), defined in the Equation \ref{eq:pseudo-calculating-pixel-class}.

% \begin{multline}
\begin{equation}
\label{eq:pseudo-calculating-pixel-class}
    C(nr, nc) \equiv (0 \leq nr < H) \wedge (0 \leq nc < W) \wedge (\neg V_{nr, nc}) \wedge (|P_{nr, nc} - P_{ref}| \leq \tau)
\end{equation}
% \end{multline}

Here, \(V\) is a boolean map indicating if a pixel has already been visited, and \(\tau\) is a predefined intensity similarity threshold. In essence, a pixel is added to the mask if it is within the image bounds, has not been visited, and its intensity is close to the reference seed value. The BFS naturally terminates when no more unvisited neighboring pixels satisfy this condition, which occurs when the search frontier reaches the dilated boundaries whose pixel values fall outside the similarity threshold \(\tau\).

The interior mask extraction builds on the Seeded Region Growing (SRG) paradigm formalized by \citet{Adams1994-hz}. We extend this foundation in three ways motivated by the specific structural conditions of unlabeled granular material CT data. First, to eliminate pixel-processing order dependencies identified by \citet{Mehnert1997-wr}, the aggregation phase is structured as a BFS traversal using a first-in, first-out queue, allowing the frontier to expand symmetrically around the reference seed value. Second, unlike methods such as \citet{Cheng2020-sb} that expand pre-specified seed regions using globally defined phase thresholds, our seed identification derives a local reference value automatically via raster scan, requiring no prior knowledge of phase count or intensity ranges. This is essential for unlabeled monochromatic CT data where intra-particle intensity profiles are heterogeneous. Third, rather than assuming standard connected neighborhoods \citep{Hojjatoleslami1998-md}, BFS propagation is deliberately restricted to a $5\times5$ neighborhood with a $2$-pixel reach. This spatial constraint encodes the physical assumption that adjacent particles are separated by gaps narrower than an unrestricted frontier would cross, providing resistance to boundary leakage across the narrow, high-contrast inter-particle gaps characteristic of granular material CT scans.

\subsubsection*{Step 4 - Refinement}

\begin{figure}[!htbp]
  \begin{subfigure}[b]{0.49\linewidth}
    \includegraphics[width=\linewidth]{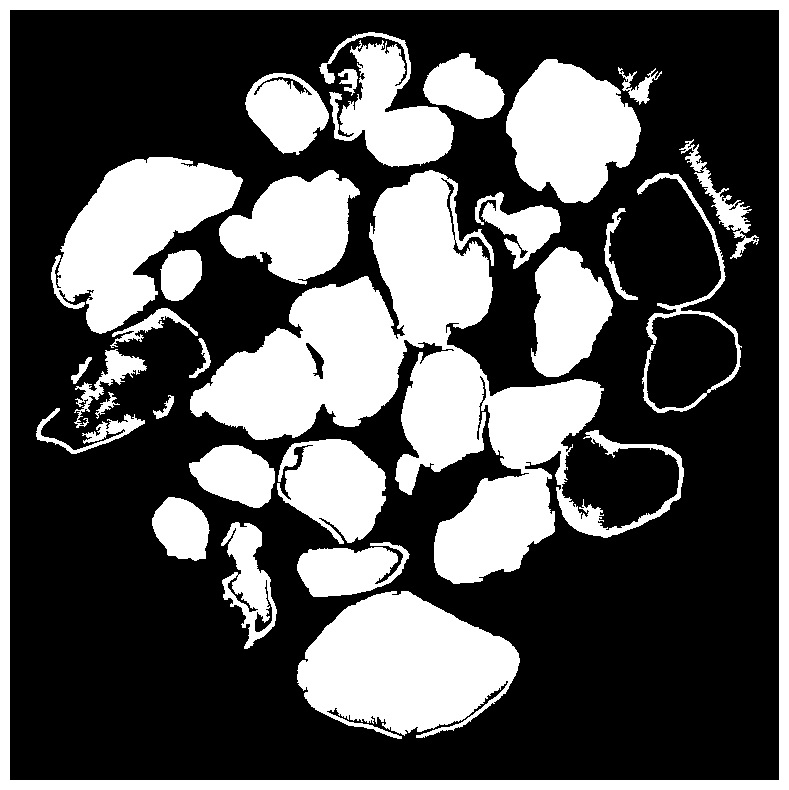}
    \caption{Masks after region growing.}
    \label{fig:pseudo-region-growing-mask}
  \end{subfigure}
  \hfill
  \begin{subfigure}[b]{0.49\linewidth}
    \includegraphics[width=\linewidth]{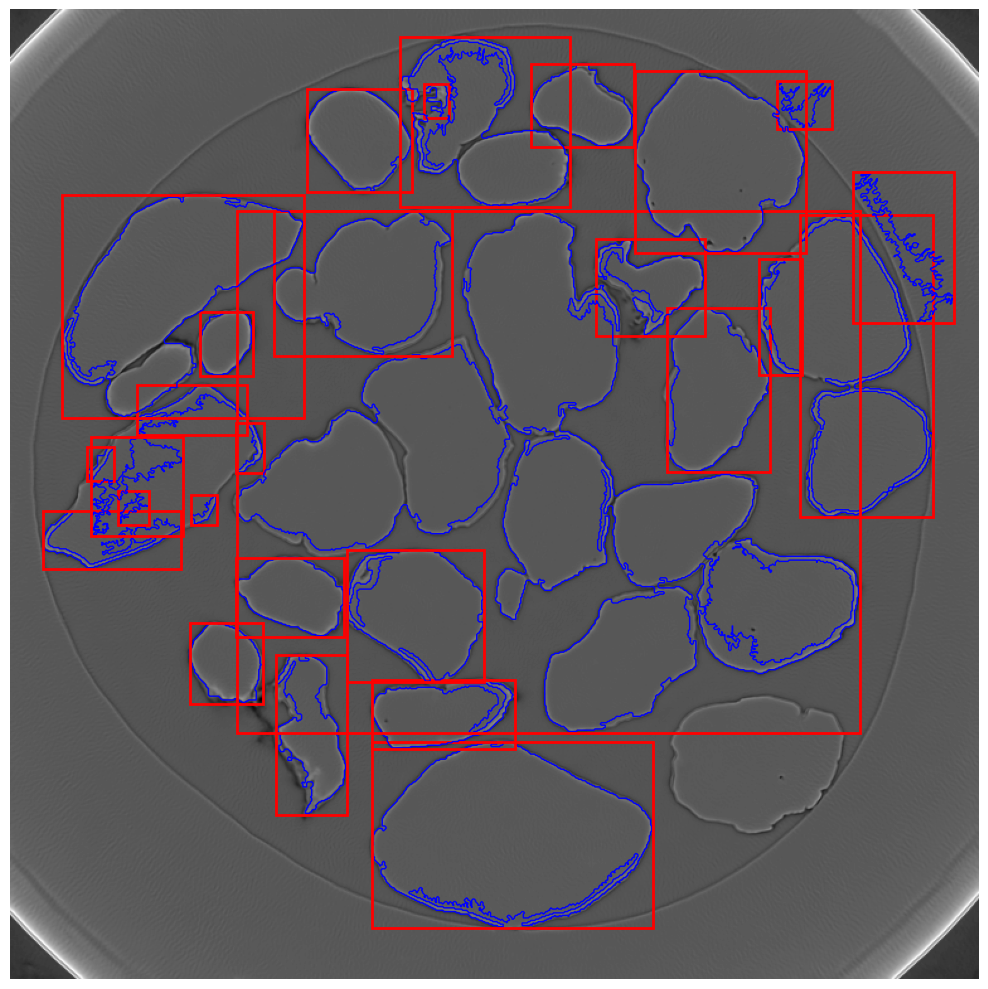}
    \caption{Pseudo-labels.}
    \label{fig:pseudo-pseudo-labels}
  \end{subfigure}
  \caption{Masks and pseudo-labels for image segmentation models.}
\end{figure}

The final step of the pipeline involves refining the generated masks and converting them into a standard annotation format suitable for training downstream models. The masks produced by the region-growing algorithm, as shown in Fig. \ref{fig:pseudo-region-growing-mask}, are nearly complete but may include noise around the region borders, primarily isolated pixels or small protrusions left where the BFS search partially leaked past a low-contrast boundary. To refine the generated masks, we apply \texttt{remove\_small\_objects} and \texttt{remove\_small\_holes} from \texttt{skimage.morphology}, both with an area threshold of $150$ pixels, removing spurious foreground fragments and filling small interior gaps below this size. We adopt the same $150$ pixel threshold used in the matched-scale Watershed comparison reported in Section \ref{sec:pseudo-quantitative-evaluation}, ensuring both methods are evaluated under an identical noise-removal criterion. Following this refinement, connected-component analysis is performed a second time to establish definitive object labels because the $150px$ removal step can split a mask into disconnected fragments. Then each resulting connected region is treated as one final object instance.

Finally, these pixel-based masks are converted into vector-based polygonal annotations by tracing the contours of each labeled region. As depicted in Fig. \ref{fig:pseudo-pseudo-labels}, the resulting polygons constitute the pseudo-label dataset. These annotations provide the necessary foundation for training subsequent object detection and instance segmentation models.

To facilitate reproducibility and encourage further research, the complete source code for the proposed framework is publicly available\footnote{https://github.com/DATALab-NIU/Model-free-pseudo-label-generation}.

\section{Experiment} \label{sec:pseudo-experiment}
To validate the efficacy of our model-free pseudo-labeling framework, we performed both qualitative and quantitative assessments of the generated labels.

The qualitative analysis focused on a downstream segmentation task to evaluate the practical utility of the pseudo-labels. We trained two separate YOLOv11 models \citep{Khanam2024-hp}, where one model utilized pseudo-labels from our proposed method and the other served as a benchmark trained on labels from an established DL approach. By comparing the final performance of these models, we evaluated the effectiveness of our computationally efficient approach against more resource-intensive alternatives.

For the quantitative validation, we evaluated the pseudo-labels generated on a baseline dataset containing pre-defined ground truth as described in Section \ref{sec:pseudo-validation-dataset}. We generated pseudo-labels for the regions of interest from this dataset and objectively measured their accuracy against the established ground-truth annotations.

\subsection{Experimental Setup} \label{sec:pseudo-experimental-setup}
All experiments were conducted on a CUDa-enabled workstation with an Intel Core $i9$ $3.2GHz$ processor, a single $12GB$ GPU, $64GB$ of RAM, and implemented in Python $3.11$.

\subsection{Qualitative Evaluation} \label{sec:pseudo-qualitative-evaluation}
\begin{figure}[!tbp]
  \begin{subfigure}[b]{0.49\linewidth}
    \includegraphics[width=\linewidth]{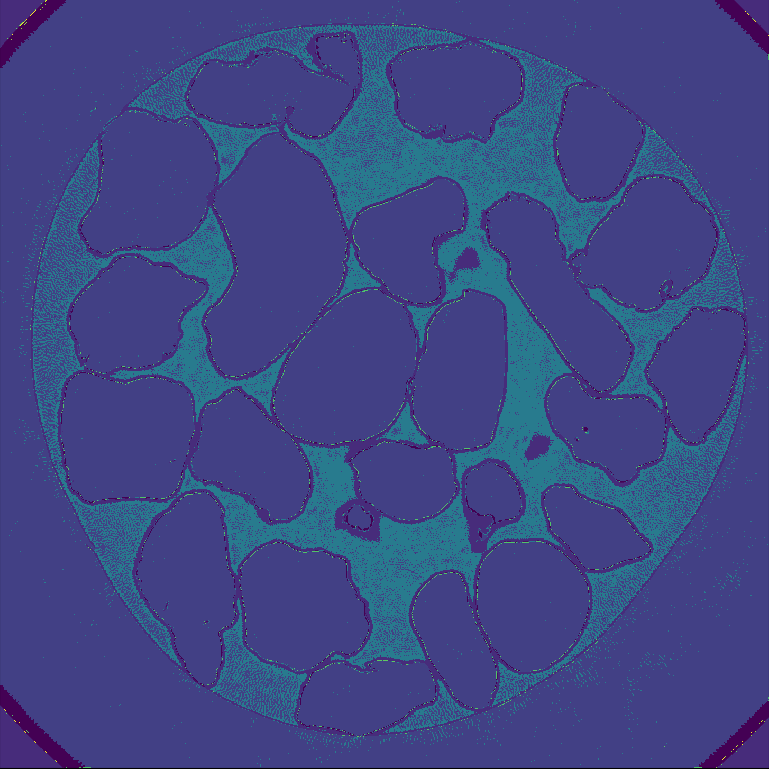}
    \caption{Segmentation using unsupervised W-Net model.}
    \label{fig:w-net-segmentation}
  \end{subfigure}
  \hfill
  \begin{subfigure}[b]{0.49\linewidth}
    \includegraphics[width=\linewidth]{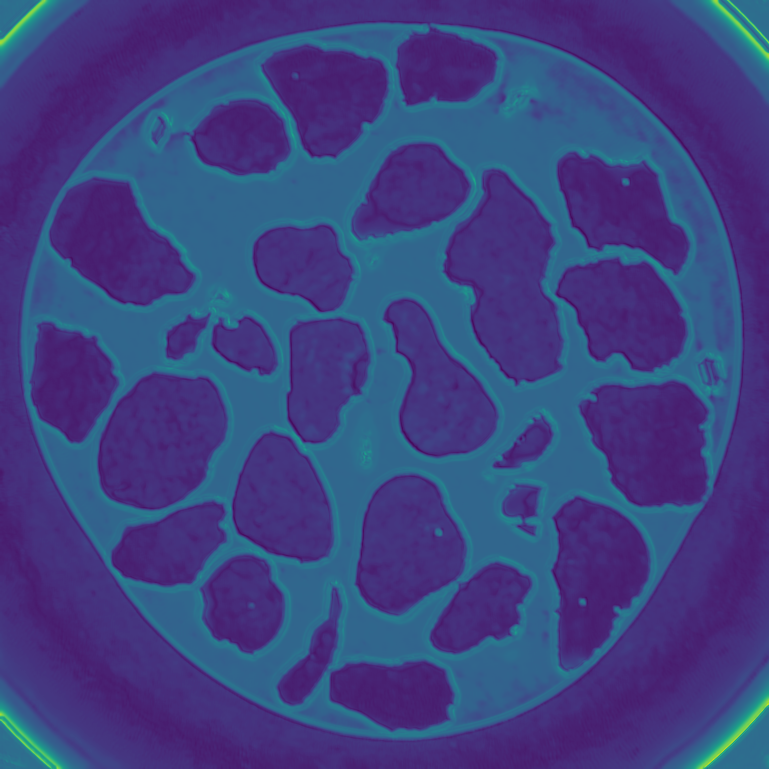}
    \caption{Segmentation using unsupervised CUTS model.}
    \label{fig:cuts-segmentation}
  \end{subfigure}
  \caption{Label generation attempts from W-Net and CUTS models.}
\end{figure}

To establish a robust benchmark, we first generate a comparative set of labels using state-of-the-art unsupervised DL-based segmentation models. Our initial candidates were the W-Net \citep{Xia2017-rd} and CUTS \citep{Liu2024-kk} models. However, both models proved unsuitable for the grayscale CT images. The W-Net model, for instance, failed to adequately distinguish sand grains from the background. While it could identify some object boundaries, the resulting segmentations were extremely noisy and unusable, as shown in Fig. \ref{fig:w-net-segmentation}. The CUTS model offered a marginal improvement by differentiating the grain interiors from their immediate surroundings. However, its performance remained insufficient, producing indistinct boundaries and noisy masks with significant misclassifications, as shown in Fig. \ref{fig:cuts-segmentation}. The failure of unsupervised DL methods may stem from their lack of specific inductive biases necessary to handle the low-contrast, heterogeneous intensity profiles common in CT images.

Then we turned to modern foundation models to generate labels as a benchmark. We generated pseudo-labels using the zero-shot Segment Anything Model (SAM) \citep{Kirillov2023-dh}, specifically the SAM $2.1$ base model with $80.9million$ parameters\footnote{https://docs.ultralytics.com/models/sam-2/}. We refined the high-quality masks generated by SAM by filtering out objects outside a specific size threshold to remove background noise. Given its substantially better performance on object-level detection, SAM was selected as the primary benchmark for further qualitative analysis.

This process resulted in two distinct pseudo-label datasets, one from our model-free method and one from SAM. These were then used to train separate YOLOv11 models for object detection and instance segmentation, forming the basis of our comparative validation.

\subsubsection*{Comparison of Pseudo-Labels} \label{sec:pseudo-pseudo-labels-comparison}
A direct comparison of the pseudo-labels generated by our model-free approach and the zero-shot SAM reveals important trade-offs in label quality, explainability, and computational cost.

\begin{figure}[!htb]
  \begin{subfigure}[b]{0.49\linewidth}
    \includegraphics[width=\linewidth]{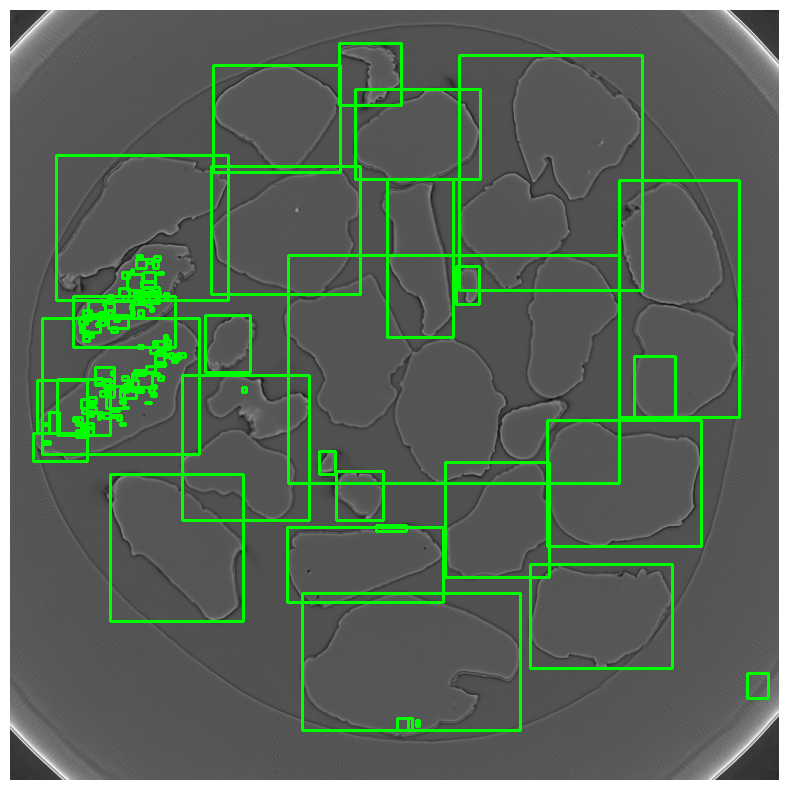}
    \caption{Model-free pseudo-labels.}
    \label{fig:pseudo-model-free-pseudo-labels-detection}
  \end{subfigure}
  \hfill
  \begin{subfigure}[b]{0.49\linewidth}
    \includegraphics[width=\linewidth]{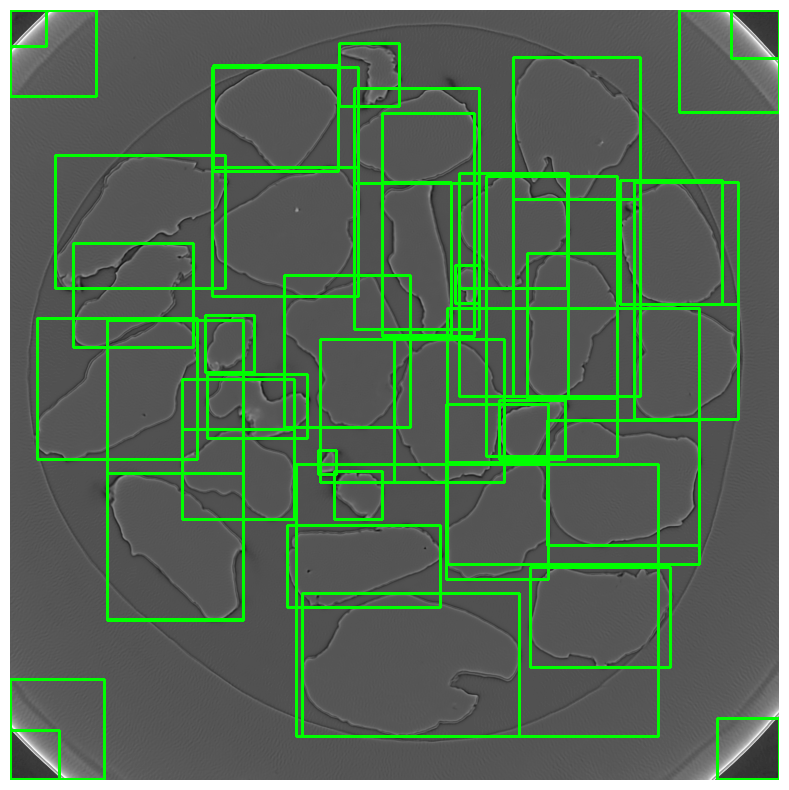}
    \caption{SAM pseudo-labels.}
    \label{fig:pseudo-sam-pseudo-labels-detection}
  \end{subfigure}
  \caption{Comparison of object detection pseudo-labels generated using model-free approach and SAM.}
\end{figure}

\begin{figure}[!hbt]
  \begin{subfigure}[b]{0.49\linewidth}
    \includegraphics[width=\linewidth]{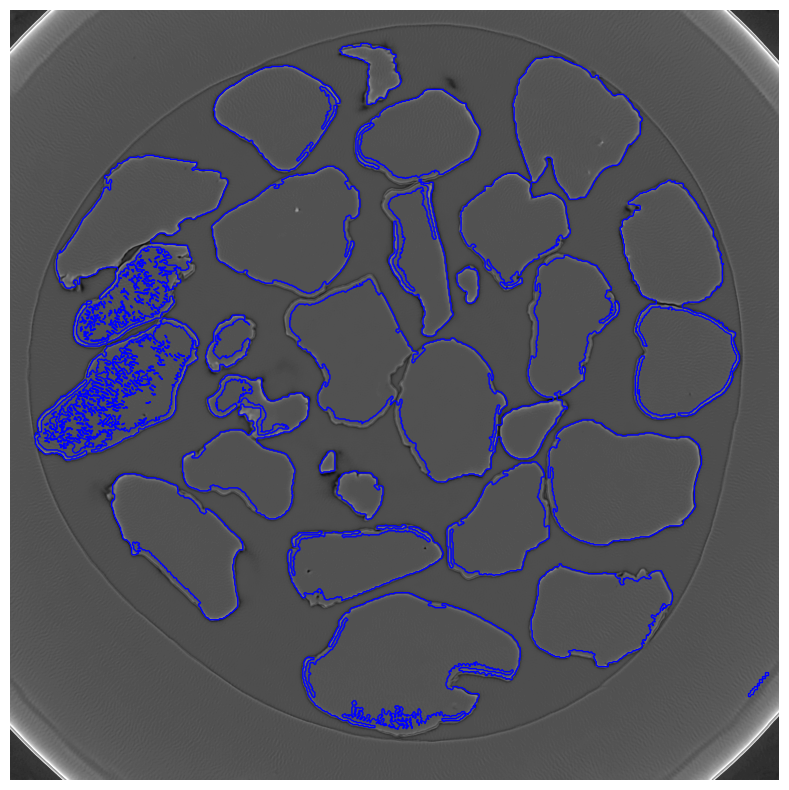}
    \caption{Model-free pseudo-labels.}
    \label{fig:pseudo-model-free-pseudo-labels-segmentation}
  \end{subfigure}
  \hfill
  \begin{subfigure}[b]{0.49\linewidth}
    \includegraphics[width=\linewidth]{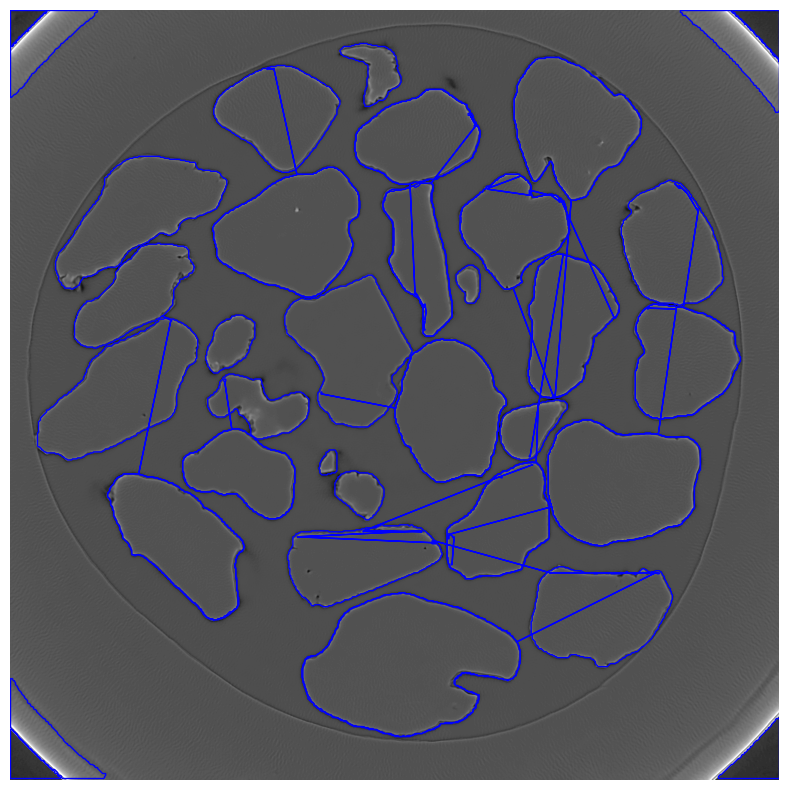}
    \caption{SAM pseudo-labels.}
    \label{fig:pseudo-sam-pseudo-labels-segmentation}
  \end{subfigure}
  \caption{Comparison of image segmentation pseudo-labels generated using model-free approach and SAM.}
\end{figure}

A visual comparison of the generated bounding boxes is presented in Figures Fig. \ref{fig:pseudo-model-free-pseudo-labels-detection} and Fig. \ref{fig:pseudo-sam-pseudo-labels-detection}, while the instance segmentation results are shown in Figures Fig. \ref{fig:pseudo-model-free-pseudo-labels-segmentation} and Fig. \ref{fig:pseudo-sam-pseudo-labels-segmentation}. Both methods produce qualitatively comparable pseudo-labels, but they exhibit distinct noise characteristics. The noise exhibited in our model-free labels can be directly attributed to specific phenomena like motion artifacts, such as double edging or shifts in grayscale values between image slices. A key advantage of our transparent framework is that this noise can be systematically mitigated by tuning the parameters in Equations \ref{eq:pseudo-delta-set} and \ref{eq:pseudo-calculating-pixel-class} to adapt the algorithm to specific dataset characteristics. In contrast, the noise in the SAM-generated labels, such as excessively overlapping bounding boxes in object detection labels, evident in Fig. \ref{fig:pseudo-sam-pseudo-labels-detection}, or the artifacts like sharp, unnatural jumps and overlapping lines in segmentation labels, evident in Fig. \ref{fig:pseudo-sam-pseudo-labels-segmentation}, is difficult to explain or correct due to the black box nature of the model.

The computational performance difference is also stark, as detailed in Table \ref{tab:pseudo-label-generation-time}. Our model-free approach averaged $15$ seconds per image on a CPU, while the SAM-based method required $28$ seconds per image, even while leveraging a $12GB$ GPU acceleration. This highlights the significant computational efficiency and accessibility of our proposed method.

\begin{table}[!hbt]
    \footnotesize
    \centering
    \caption{Comparison of resources for label generation showing the faster throughput of the model-free approach without GPU resources.}
    \begin{tabular}{lccccr}
        \hline
         Method & Parameters(Mi) & CPU & GPU & RAM & Time(sec) \\
        \hline
         Model-free & $-$ & $\checkmark$ & $\times$ & $\checkmark$ & $15/im$ \\
         \hline
         SAM & $80.9$ & $\checkmark$ & $\checkmark$ & $\checkmark$ & $28/im$ \\
         \hline
    \end{tabular}
    \label{tab:pseudo-label-generation-time}
\end{table}

\subsubsection*{Object Detection}\label{sec:pseudo-object-detection}
To evaluate our pseudo-labels, we trained the lightweight YOLOv11n\footnote{https://docs.ultralytics.com/tasks/detect/} object detection model, which has $2.6 million$ parameters, on two datasets, one using our model-free pseudo-labels and one using labels from SAM. The dataset was partitioned into $762$ training, $191$ validation, and $239$ test images.

\begin{figure}[!htb]
  \begin{subfigure}[b]{0.49\linewidth}
    \includegraphics[width=\linewidth]{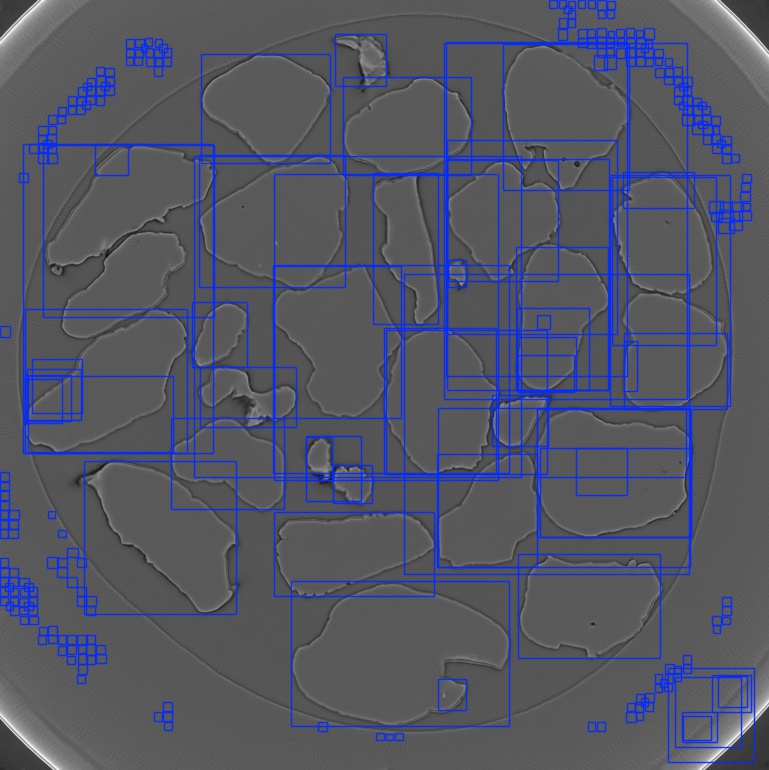}
    \caption{Model-free pseudo-labels.}
    \label{fig:pseudo-yolo-object-detection-model-free}
  \end{subfigure}
  \hfill
  \begin{subfigure}[b]{0.49\linewidth}
    \includegraphics[width=\linewidth]{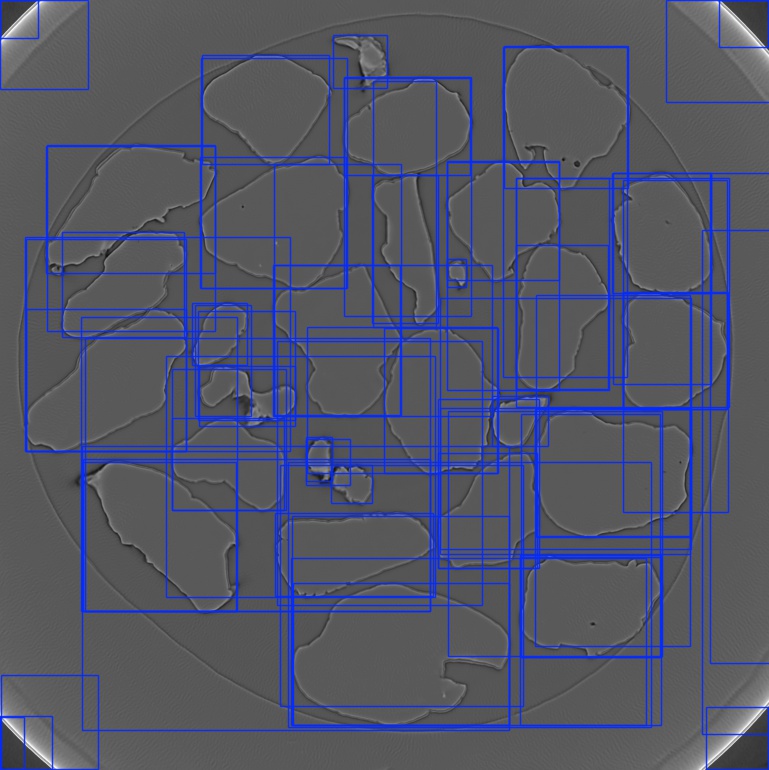}
    \caption{SAM pseudo-labels.}
    \label{fig:pseudo-yolo-object-detection-sam}
  \end{subfigure}
  \caption{Comparison of object detection models trained with model-free pseudo-labels and SAM-based pseudo-labels}
\end{figure}

A qualitative comparison shows both models perform comparably. As shown in Fig. \ref{fig:pseudo-yolo-object-detection-model-free}, the model trained with our model-free labels detected most sand grains but showed some noise and over-segmentation, which can be directly attributed to the noise present in its training data. The model trained with SAM generated labels also detected the grains effectively but produced more redundant and heavily overlapping bounding boxes, as shown in Fig. \ref{fig:pseudo-yolo-object-detection-sam}. This suggests that while both labeling methods are effective for training, the specific type of noise inherent in each dataset directly influences the characteristic errors of the final model.

\subsubsection*{Instance Segmentation}\label{sec:pseudo-instance-segmentation}
Following the object detection experiments, we evaluated the pseudo-labels on the more challenging task of instance segmentation using a lightweight YOLOv11n-seg\footnote{https://docs.ultralytics.com/tasks/segment/} model, which has $2.9 million$ parameters. Using the same data partitions, we trained one model on our model-free labels and another on labels from SAM.

\begin{figure}[!hbt]
  \begin{subfigure}[b]{0.49\linewidth}
    \includegraphics[width=\linewidth]{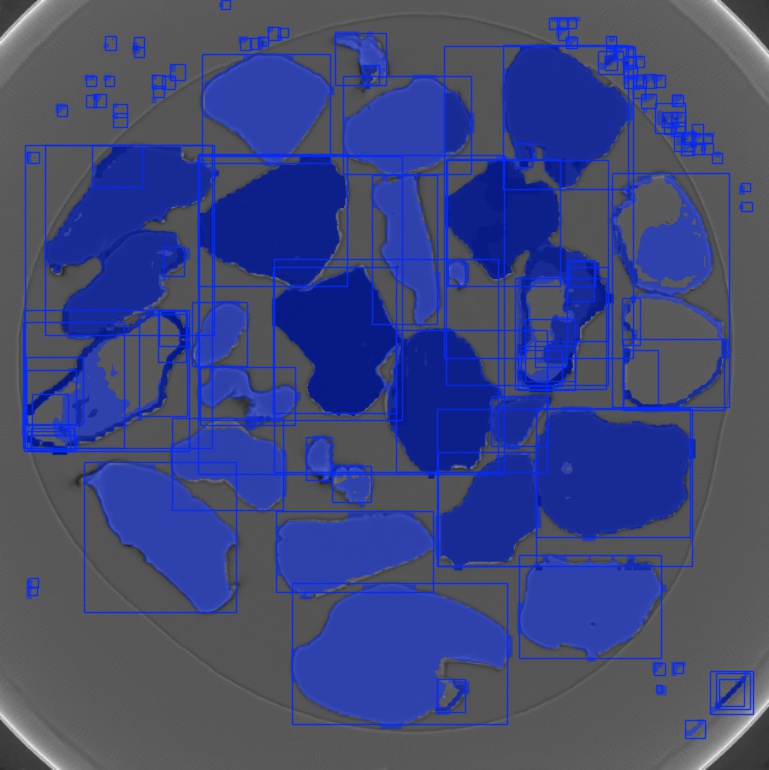}
    \caption{Model-free pseudo-labels.}
    \label{fig:pseudo-yolo-segmentation-model-free}
  \end{subfigure}
  \hfill
  \begin{subfigure}[b]{0.49\linewidth}
    \includegraphics[width=\linewidth]{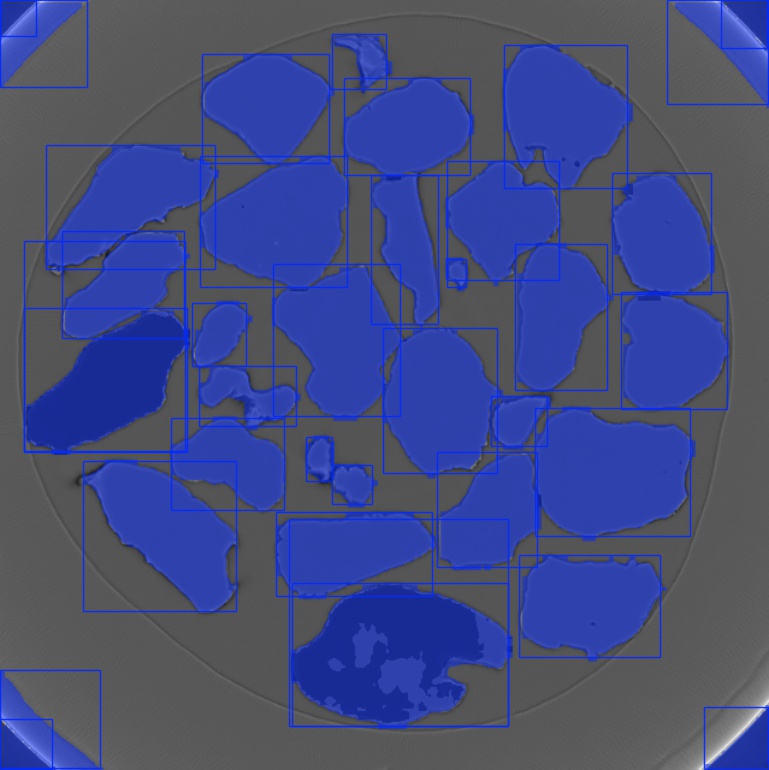}
    \caption{SAM pseudo-labels.}
    \label{fig:pseudo-yolo-segmentation-sam}
  \end{subfigure}
  \caption{Comparison of instance segmentation models trained with model-free pseudo-labels and SAM-based pseudo-labels}
\end{figure}

The qualitative results are shown in Figures Fig. \ref{fig:pseudo-yolo-segmentation-model-free} and Fig. \ref{fig:pseudo-yolo-segmentation-sam}. The model trained with SAM generated labels achieved more complete segmentations. However, the model trained with our labels delivered competitive performance, accurately capturing the shape of most grains. Its primary error, unsegmented voids within some grains, directly corresponds to small gaps present in our pseudo-labels.

A critical advantage of our transparent framework is that errors are directly traceable to their source and can be systematically mitigated by tuning the parameters in Equations \ref{eq:pseudo-delta-set} and \ref{eq:pseudo-calculating-pixel-class}. In contrast, correcting the less predictable failure modes of the black-box SAM model is non-trivial. This highlights a fundamental trade-off. While the DL-based approach may achieve higher initial accuracy, our model-free method is significantly more computationally efficient, offers superior explainability, and provides a more direct pathway for iterative refinement.

\subsection{Quantitative Evaluation} \label{sec:pseudo-quantitative-evaluation}
\begin{figure}[!hbt]
  \begin{subfigure}[b]{0.24\linewidth}
    \includegraphics[width=\linewidth]{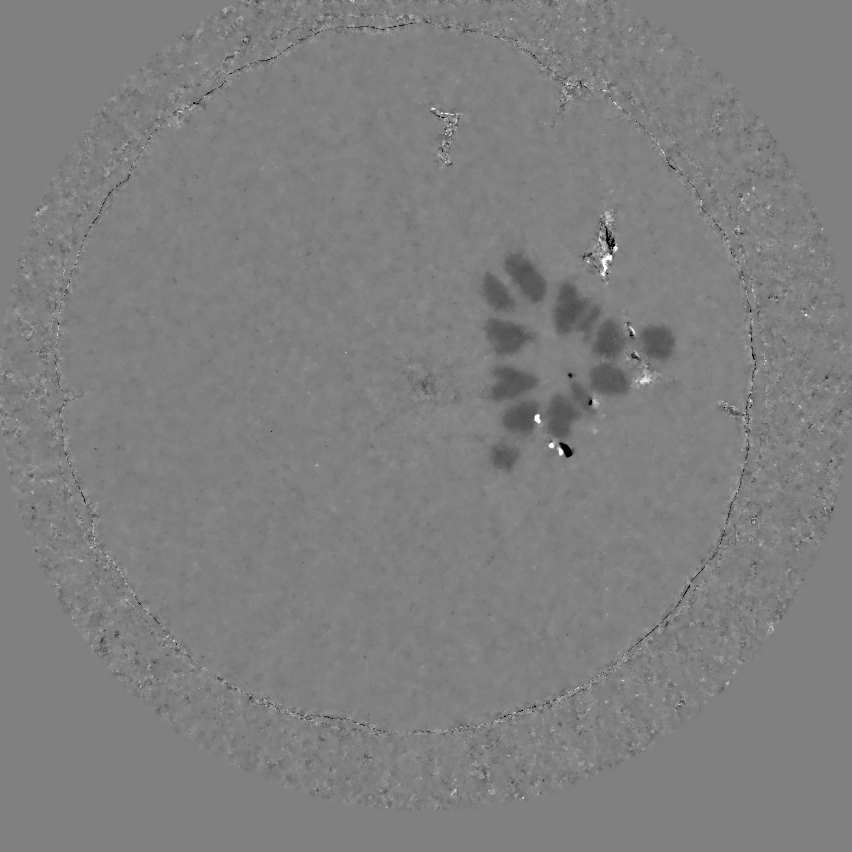}
    \caption{Original scan image of a dendrite.}
    \label{fig:pseudo-dendrite-original}
  \end{subfigure}
  % \hfill
  % \begin{subfigure}[b]{0.24\linewidth}
  %   \includegraphics[width=\linewidth]{dendrite_gt}
  %   \caption{Ground truth label of dendrite.}
  %   \label{fig:pseudo-dendrite-gt}
  % \end{subfigure}
  \hfill
  \begin{subfigure}[b]{0.24\linewidth}
    \includegraphics[width=\linewidth]{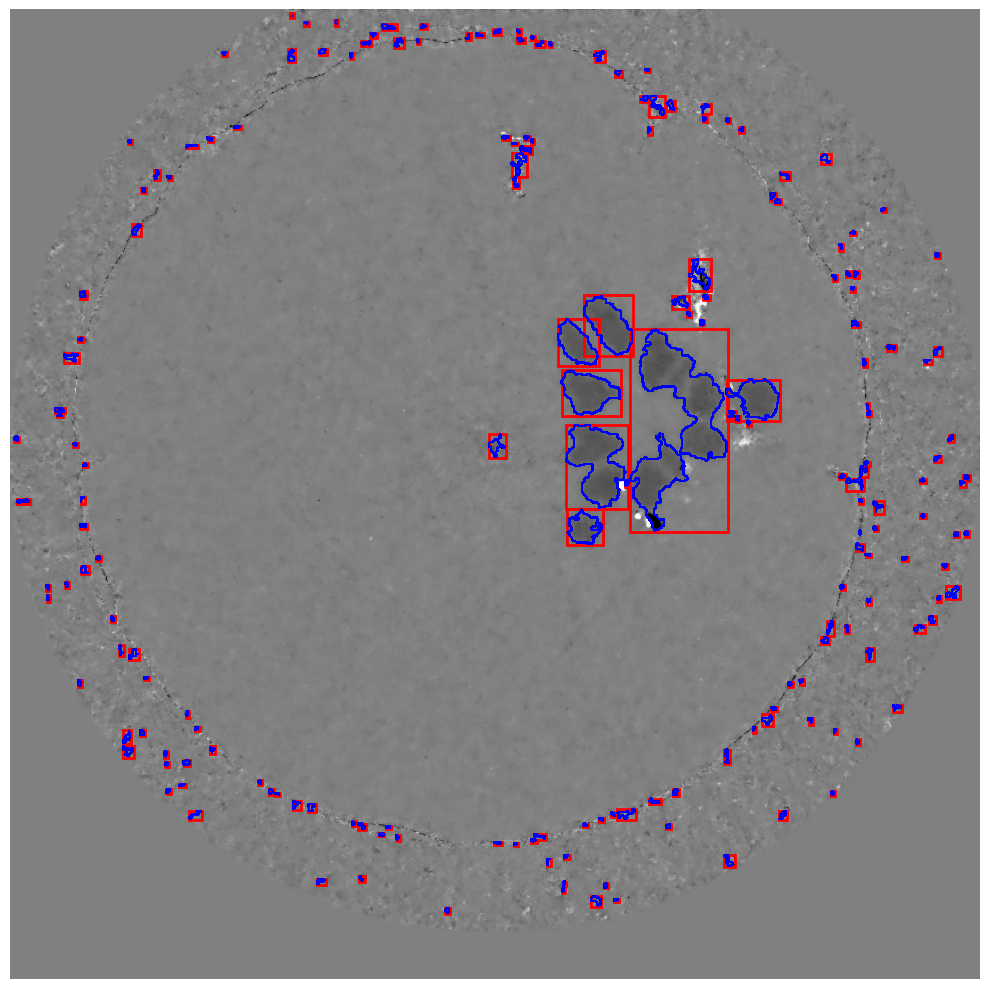}
    \caption{Pseudo-labels using Watershed.}
    \label{fig:pseudo-dendrite-watershed}
  \end{subfigure}
  \hfill
  \begin{subfigure}[b]{0.24\linewidth}
    \includegraphics[width=\linewidth]{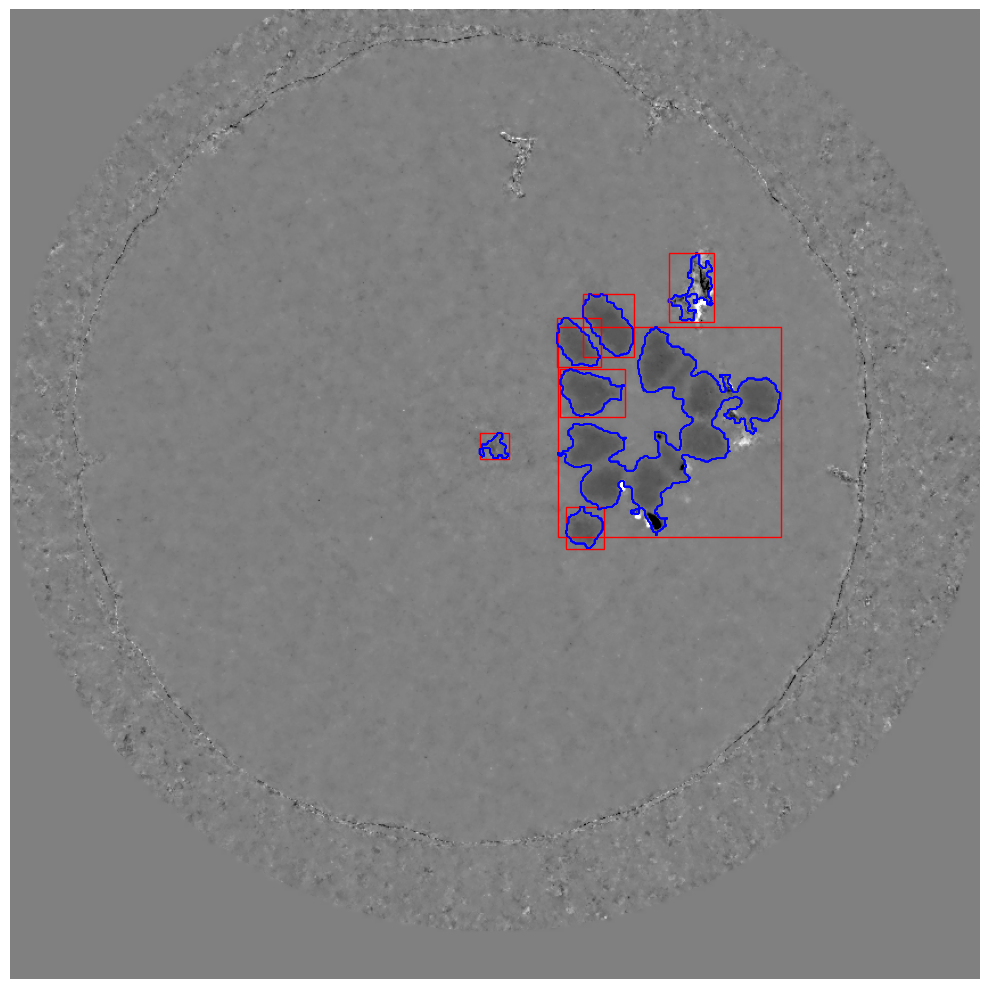}
    \caption{Pseudo-labels using our algorithm.}
    \label{fig:pseudo-dendrite-pseudo-label}
  \end{subfigure}
  \hfill
  \begin{subfigure}[b]{0.24\linewidth}
    \includegraphics[width=\linewidth]{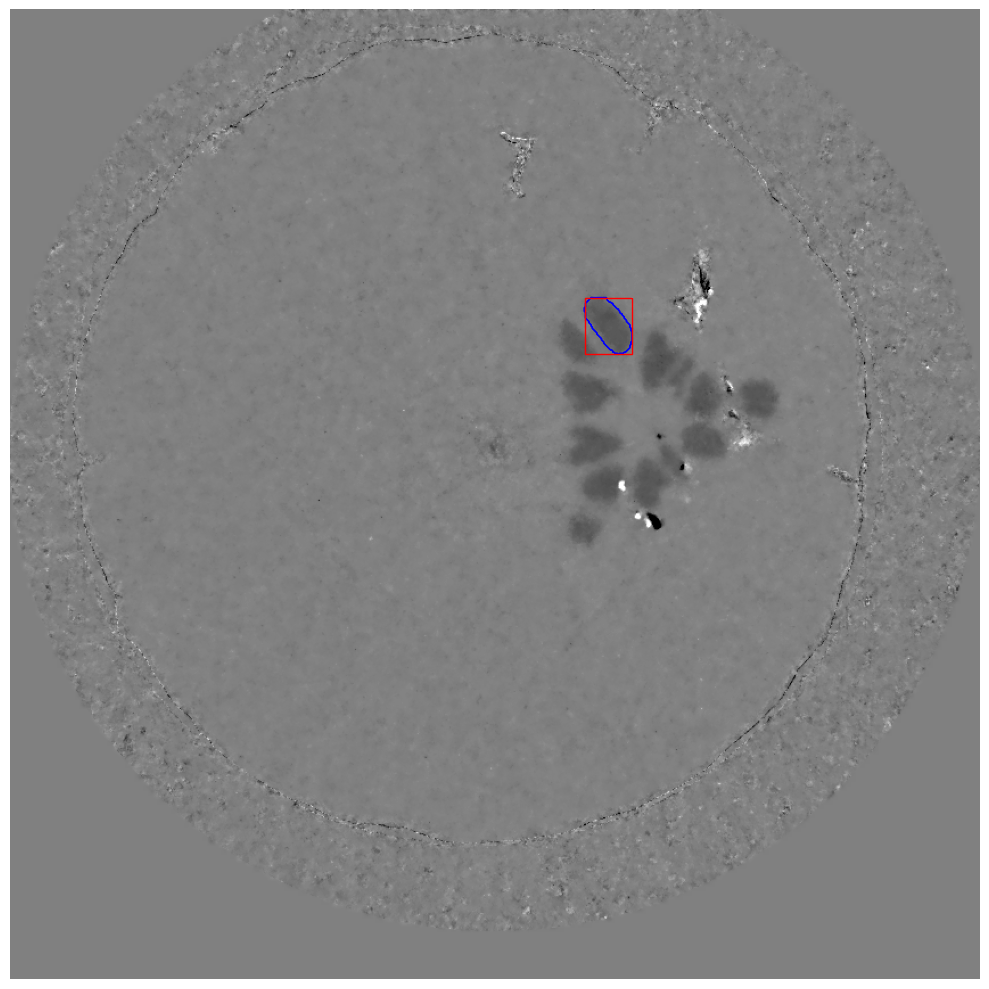}
    \caption{Pseudo-labels generated by SAM.}
    \label{fig:pseudo-dendrite-pseudo-label-sam}
  \end{subfigure}
  \caption{Visual comparison of pseudo-labels generated by our algorithm (c) with the Watershed approach (b), and zero-shot SAM (d).}
  \label{fig:pseudo-dendrite-comparison}
\end{figure}

For the quantitative analysis, we generated pseudo-labels from images in the Dendrites dataset to compare our algorithm against the established ground truth. We implemented the proposed algorithm in the established experimental environment and compared its results with Watershed algorithm performance and SAM's zero-shot performance. Fig. \ref{fig:pseudo-dendrite-comparison} presents a visual comparison where Fig. \ref{fig:pseudo-dendrite-original} shows the original scan image, Figures Fig. \ref{fig:pseudo-dendrite-watershed}, Fig. \ref{fig:pseudo-dendrite-pseudo-label}, and Fig. \ref{fig:pseudo-dendrite-pseudo-label-sam} show the generated pseudo-labels using the Watershed algorithm, our model-free algorithm, and SAM, respectively.

Table \ref{tab:pseudo-pixel-results} summarizes the pixel-level performance for the Watershed algorithm, our model-free algorithm, and SAM in a zero-shot setting. For object-level performance, we report the Mean Precision $@50$ ($AP@50$), Mean Recall $@50$ ($AR@50$), Mean Precision $@75$ ($AP@75$), and Mean Recall $@75$ ($AR@75$), calculated by averaging the precision and recall at IoU thresholds $0.5$ and $0.75$ across all images in the dataset. Table \ref{tab:pseudo-object-results} details the object-level metrics from our experiment.

At the pixel level, our model-free approach achieved an IoU of $63.19\%$ and an F1 score of $77.44\%$. In contrast, SAM produced an IoU of $10.74\%$ and an F1 score of $19.39\%$. While both methods showed high pixel accuracy scores above $98\%$ this metric is misleading due to the large proportion of background area in the CT scans. The significantly higher IoU and F1 scores of our algorithm prove its ability to capture the complex morphology of the dendrite structures more effectively than the foundation model. We noticed that while the Watershed algorithm performed as close as our proposed framework, it needed excessive fine-tuning. While, both our model-free framework and SAM was able to identify connected regions, ignoring smaller noisy areas smaller than $150px$, Watershed algorithm failed to do that. Watershed algorithm had the peak performance when considered smaller pixel areas of $10px$ and above. This indicates that Watershed algorithm, while similar in quantitative values, only detects fragmented smaller regions, which is not useful for downstream tasks like training DL-based segmentation models. This over-segmentation issue is visible in Fig. \ref{fig:pseudo-dendrite-watershed}. 

\begin{table}[!hbt]
    \footnotesize
    \centering
    \caption{Pixel-level quantitative comparison of pseudo-label generation showing high performance of the model-free approach with zero labeled data. The 3D U-Net shows a reference ceiling where a large set of labeled data is required.}
    \begin{tabular}{lccccccc}
        \hline
         Method & Accuracy & IoU & F1 \\
        \hline
         SAM & $98.63\%$ & $10.74\%$ & $19.39\%$ \\
         \hline
         Watershed @150px & $98.50\%$ & $2.86\%$ & $5.56\%$ \\
         \hline
         Watershed @10px & $99.06\%$ & $60.76\%$ & $75.59\%$ \\
         \hline
         3D U-Net \citep{James2023-dc} & $99.84\%$ & $95.56\%$ & $98.59\%$ \\
         \hline
         Model-free & \textbf{$99.12\%$} & \textbf{$63.19\%$} & \textbf{$77.44\%$} \\
         \hline
    \end{tabular}
    \label{tab:pseudo-pixel-results}
\end{table}

To further contextualize the performance of our framework, we compared our results against the supervised state-of-the-art benchmark established for this specific dendritic dataset by \citet{James2023-dc}, which we treat as a reference ceiling rather than a peer competitor given its dependence on expert-annotated training data unavailable in our zero-label setting. Their optimized $3D$ U-Net achieved a maximum IoU of $95.56\%$ through an extensive two-step hyperparameter optimization and fine-tuning process on high-end NVIDIA $A100$ GPUs. Our proposed model-free framework achieved $63.19\%$ IoU, a result that is significantly closer to the supervised ceiling than the $10.74\%$ IoU produced by the zero-shot SAM. Notably, our framework attains this performance with no training requirements and no manual labeling overhead, which \citet{James2023-dc} identified as a critical bottleneck requiring approximately $20$ minutes of expert labor per image slice. These findings establish our method as a viable bridge for rapid characterization in scientific and industrial environments where the computational and labor costs of supervised deep learning are prohibitive.

Object-level metrics derived from the Dendrites dataset further highlight the trade-offs between the methods. On these complex, hyperbranched structures, SAM demonstrated a precision of $100\%$ at both $50\%$ and $75\%$ overlap thresholds, indicating high confidence on a strictly limited subset of features. However, its recall remained exceptionally low at $8.33\%$. This low recall reveals a distinct failure mode dictated by material morphology. While foundation models tend to hallucinate features in tightly packed, noisy environments, they become overly conservative when faced with the diffuse, continuous solid-liquid boundaries characteristic of dendritic structures. Consequently, SAM failed to recognize the majority of the objects in the scan. The over-segmentation issue of the Watershed algorithm is also highly evident in these metrics. The average precision and recall remain extremely low when the algorithm considers small pixel fragments, as it shatters true regions into unusable pieces. Conversely, our model-free framework achieved a recall of $33.33\%$ at the $50\%$ threshold, representing a four-fold improvement over the foundation model. While our method exhibits lower precision due to some noise artifacts, its region-growing logic successfully navigates diffuse boundaries to provide a much more comprehensive and physically representative set of annotations for downstream training

\begin{table}[!hbt]
    \footnotesize
    \centering
    \caption{Instance-level quantitative comparison of pseudo-label generation showing a reliable performance of the model-free framework.}
    \begin{tabular}{lccccccc}
        \hline
         Method & $AP@{50}$ & $AR@{50}$ & $AP@{75}$ & $AR@{75}$ \\
        \hline
         SAM & $100.00\%$ & $8.33\%$ & $100.00\%$ & $8.33\%$ \\
        \hline
         Watershed @150px & $0.00\%$ & $0.00\%$ & $0.00\%$ & $0.00\%$ \\
         \hline
         Watershed @10px & $2.63\%$ & $41.67\%$ & $2.63\%$ & $41.67\%$ \\
         \hline
         Model-free & \textbf{$57.14\%$} & \textbf{$33.33\%$} & \textbf{$28.57\%$} & \textbf{$16.67\%$} \\
         \hline
    \end{tabular}
    \label{tab:pseudo-object-results}
\end{table}

\subsection{Microstructural Quantification} \label{sec:microstructure-quant}
\begin{figure}[!hbt]
  \begin{subfigure}[b]{0.48\linewidth}
    \includegraphics[width=\linewidth]{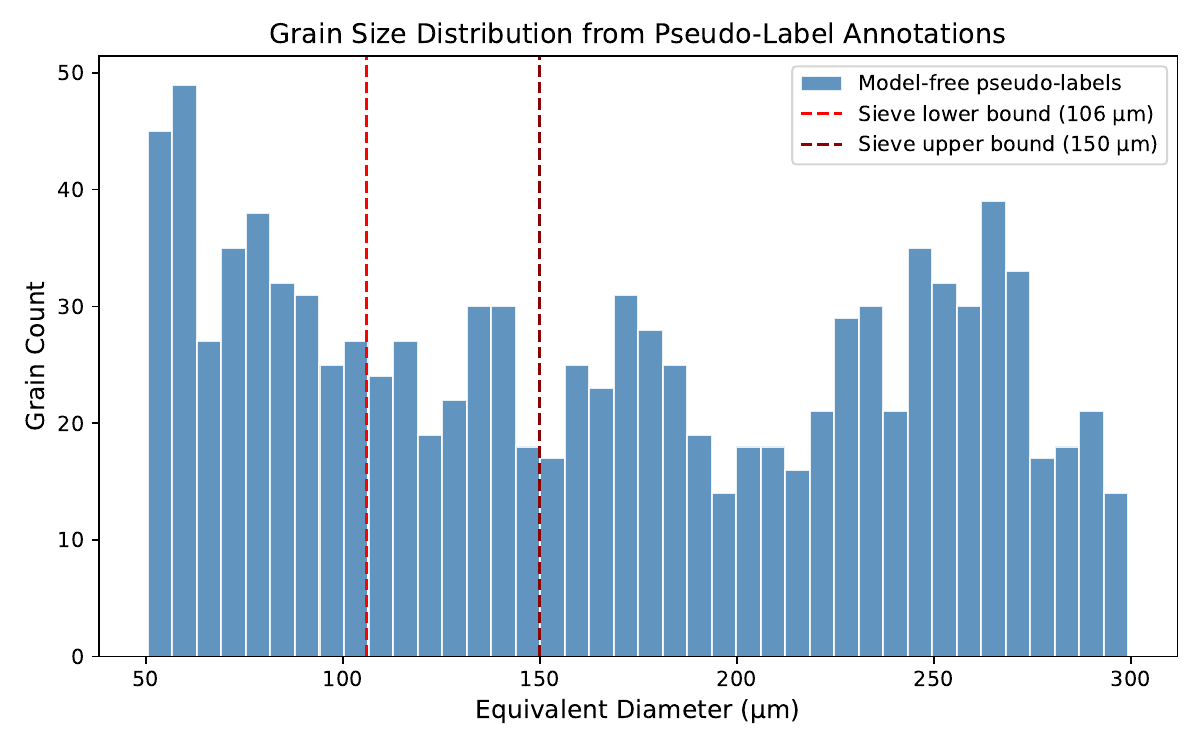}
    \caption{ECD from model-free pseudo-label.}
    \label{fig:pseudo-grain-size-distribution}
  \end{subfigure}
  \hfill
  \begin{subfigure}[b]{0.48\linewidth}
    \includegraphics[width=\linewidth]{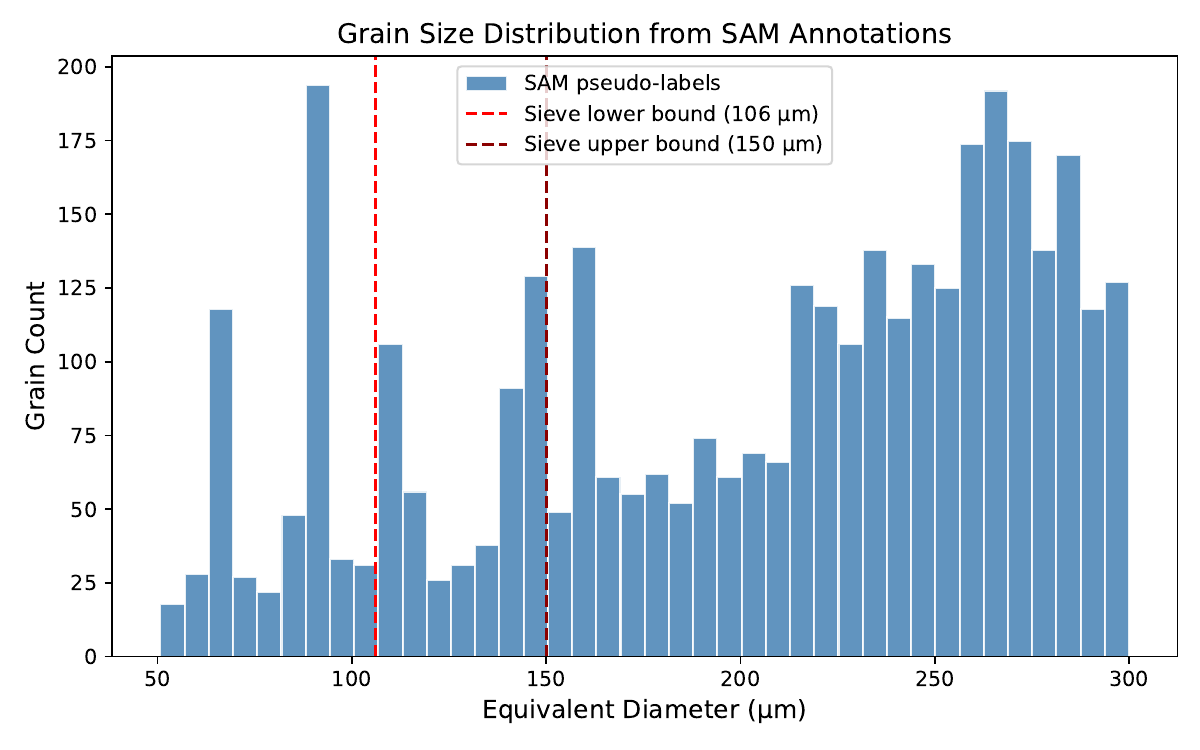}
    \caption{ECD from SAM pseudo-label.}
    \label{fig:pseudo-grain-size-distribution-sam}
  \end{subfigure}
  \caption{Comparison of ECD calculated by model-free approach and SAM at zero-shot setting.}
  \label{fig:pseudo-grain-size}
\end{figure}

To ensure statistical robustness and validate the utility of the framework across a larger volume, we performed a quantitative microstructural characterization on a batch of $100$ randomly selected slices from the primary dataset (\ref{sec:pseudo-primary-dataset}). The dataset was comprised from F-75 Ottawa sand specimen with grain sizes between U.S. sieve $\#100 (150 {\mu}m)$ and $\#140 (106 {\mu}m)$ \citep{Imseeh2020-az}.  Visual inspection confirmed a physical density of approximately $25$ grains per slice. Each polygonal annotation was converted into an equivalent circular diameter (ECD) by calculating the area and applying the $1.17 {\mu}m/pixel$ spatial resolution. To systematically remove non-physical noise and artifacts from the statistics, we applied edge-exclusion filtering and established strict ECD cut-off boundaries at $50 {\mu}m$ and $300 {\mu}m$. Following this filtration, our proposed model-free framework successfully extracted $1,053$ valid, complete grains. This filtered population exhibited a mean ECD of $166.1{\mu}m$ (SD $75.1{\mu}m$) and a median of $163.8{\mu}m$. This median closely tracks the physical $150{\mu}m$ upper sieve limit of the specimen, within the expected range of $2D$ stereological sampling variation. In contrast, the zero-shot Segment Anything Model (SAM) identified $3,640$ features with a mean diameter of $201.5{\mu}m$ (SD $70.4{\mu}m$) and a median of $219.5{\mu}m$. This distinction is evident in Figure \ref{fig:pseudo-grain-size}.

\begin{figure}[!hbt]
    \centering
    \includegraphics[width=1\linewidth]{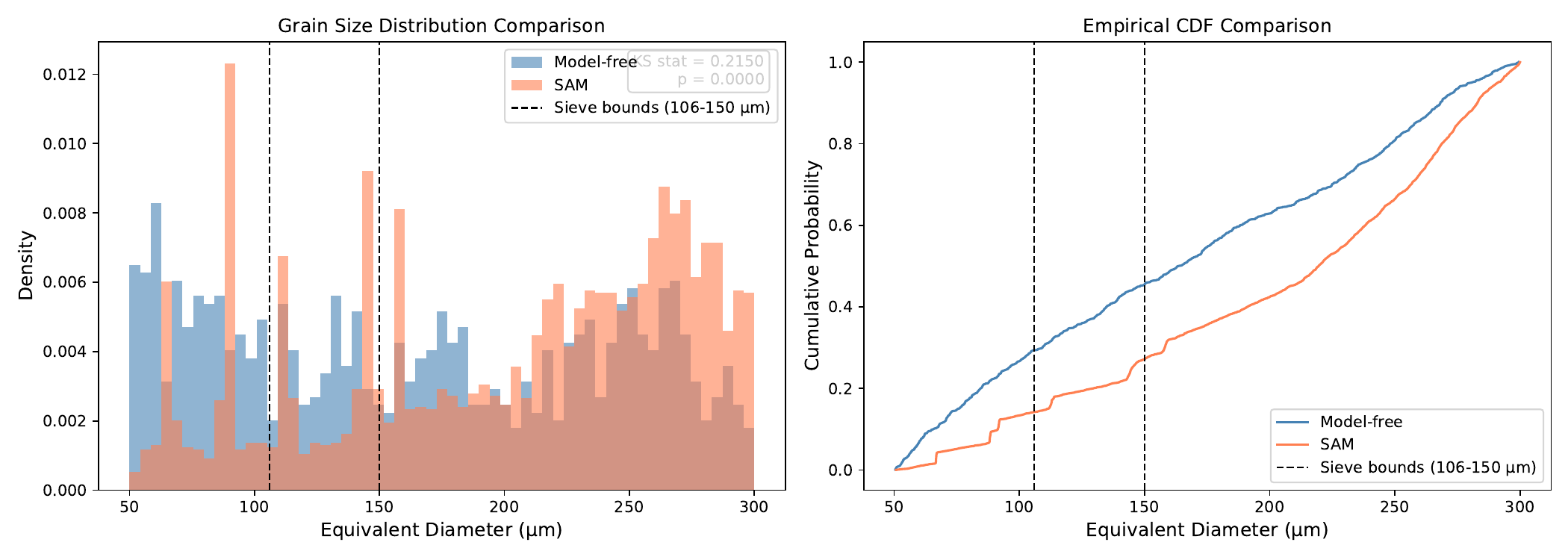}
    \caption{Statistical comparison of grain size distributions. Left, kernel density estimates of the ECD distributions for model-free and SAM pseudo-labels. Right, corresponding empirical cumulative distribution functions, with the KS statistic corresponding to their maximum vertical separation.}
    \label{fig:pseudo-ks-distribution}
\end{figure}

The statistical validity of the framework is supported by a two-sample Kolmogorov-Smirnov (KS) test ($KS = 0.215$, $p < 0.0001$), which confirmed a statistically significant divergence between the distributions. This divergence is visualized in Figure \ref{fig:pseudo-ks-distribution}, which overlays the density-normalized ECD distributions alongside their empirical cumulative distribution functions, the latter directly illustrating the maximum separation quantified by the KS statistic. This experiment indicates that the foundation model without specialized training simultaneously hallucinates instances from background noise and merges distinct, adjacent particles into artificially large structures, failing to capture a representative population of the microstructure. Conversely, our framework provides a reliable, physically grounded microstructural quantification essential for reliable grain count, morphology analysis, and ICME workflows.

\subsection{Parameter Sensitivity Analysis} \label{sec:pseudo-sensitivity-analysis}

To evaluate the stability and robustness of the proposed framework, we analyzed the effect of varying the intensity similarity threshold $\tau$ on the validation dataset (\ref{sec:pseudo-validation-dataset}). The parameter was varied incrementally from $\tau-4$ to $\tau+4$ around its baseline operational value, and the resulting pixel-level and instance-level metrics are summarized in Table~\ref{tab:pseudo-parameter-sensitivity}.

\begin{table}[!htbp]
    \footnotesize
    \centering
    \caption{Framework performance across variations of the intensity similarity threshold ($\tau$).}
    \label{tab:pseudo-parameter-sensitivity}
    \begin{tabular}{lrrrrrrrr}
        \hline
        Threshold & Prec. & Rec. & F1 & AP@50 & AR@50 & AP@75 & AR@75 \\ 
        \hline
        $\tau - 4$  & $63.40\%$ & $99.99\%$ & $77.60\%$ & $57.14\%$ & $33.33\%$ & $28.57\%$ & $16.67\%$ \\
        $\tau - 3$ & $63.40\%$ & $99.99\%$ & $77.60\%$ & $57.14\%$ & $33.33\%$ & $28.57\%$ & $16.67\%$ \\
        $\tau - 2$  & $63.40\%$ & $99.99\%$ & $77.60\%$ & $57.14\%$ & $33.33\%$ & $28.57\%$ & $16.67\%$ \\
        $\tau - 1$  & $63.40\%$ & $99.99\%$ & $77.60\%$ & $57.14\%$ & $33.33\%$ & $28.57\%$ & $16.67\%$ \\
        \textbf{$\tau + 0$} & \textbf{$63.19\%$} & \textbf{$99.99\%$} & \textbf{$77.44\%$} & \textbf{$57.14\%$} & \textbf{$33.33\%$} & \textbf{$28.57\%$} & \textbf{$16.67\%$} \\
        $\tau + 1$ & $6.31\%$  & $99.99\%$ & $11.87\%$ & $6.25\%$  & $16.67\%$ & $0.00\%$  & $0.00\%$  \\
        $\tau + 2$ & $3.42\%$  & $100.00\%$ & $6.62\%$ & $0.00\%$ & $0.00\%$ & $0.00\%$ & $0.00\%$  \\
        $\tau + 3$ & $0.00\%$  & $0.00\%$  & $0.00\%$  & $0.00\%$  & $0.00\%$  & $0.00\%$  & $0.00\%$  \\
        $\tau + 4$ & $0.00\%$  & $0.00\%$  & $0.00\%$  & $0.00\%$  & $0.00\%$  & $0.00\%$  & $0.00\%$  \\ 
        \hline
    \end{tabular}
\end{table}

The sensitivity analysis reveals a clear operational profile consisting of a broad stability plateau at lower threshold values and a sharp performance transition at higher values. From $\tau-4$ through $\tau-1$, all segmentation metrics remain unchanged relative to the baseline, indicating that the solid dendritic phase exhibits a sufficiently uniform internal intensity profile that the BFS-based region-growing algorithm is insensitive to moderate under-thresholding. This stability over a range of four threshold units suggests that the framework does not require precise threshold calibration within this regime, which is a practically useful property for deployment on novel datasets of similar material type.

Exceeding the baseline threshold, however, produces a rapid and predictable deterioration in performance. At $\tau+1$ and $\tau+2$, pixel-level recall remains near unity while precision drops below $7\%$, reflecting a boundary leakage condition in which the permissive threshold allows the BFS frontier to propagate beyond the solid-liquid interface into the background region. At $\tau+3$ and $\tau+4$, this leakage becomes sufficiently severe that no valid object masks are produced after the refinement step, reducing all instance-level metrics to zero. Rather than indicating fragility, this sharp and predictable transition is direct evidence that $\tau$ is functioning as intended. The transition point corresponds to the actual intensity contrast between the solid and liquid phases in the image, meaning $\tau$ operates as a proxy for a real physical boundary in the material rather than an arbitrary tunable knob. A parameter disconnected from any underlying physical signal would be expected to degrade gradually or unpredictably as it increased. The sharp break instead confirms the opposite, that the algorithm faithfully tracks the true phase boundary until the moment the permitted intensity tolerance exceeds it, at which point failure is immediate rather than gradual. This also means the valid operating range for $\tau$ can be identified directly from the image histogram, without labeled data, replacing arbitrary trial-and-error tuning with a physically grounded calibration step. This transparent, physically grounded sensitivity behavior distinguishes the proposed framework from black-box deep learning approaches, where equivalent failure modes are difficult to diagnose, predict, or trace to a single underlying cause.

It should be noted that the specific plateau width and transition point are properties of the Dendrites dataset and its particular phase contrast characteristics. Datasets with lower solid-liquid intensity contrast or more heterogeneous internal intensity profiles may exhibit a narrower stability plateau, requiring more careful threshold selection. This is discussed further in Section~\ref{sec:pseudo-limitations-and-future-works}.

\section{Discussion} \label{sec:pseudo-discussion}
The practical application of semi-supervised learning in specialized domains like materials science is often limited by the high computational cost and low interpretability of DL-based pseudo-labeling. Our work demonstrates that a thoughtfully designed model-free framework offers a compelling alternative. This approach is fast and transparent while remaining resource-efficient. The experimental results provide strong evidence for this position through quantitative and qualitative assessments.

In quantitative validation, our algorithm serves as a highly competitive baseline. While the fully supervised $3D$ U-Net of \citet{James2023-dc} achieves an IoU of $95.56\%$ on the Dendrites dataset, this serves as a reference ceiling rather than a peer competitor, since this benchmark requires expert-annotated training labels that are unavailable in the resource-constrained settings this work targets. Our method achieves $63.19\%$ IoU without any labeled data, representing a substantial improvement over zero-shot SAM, with IoU of $10.74\%$, in the same annotation-free setting, and providing a viable starting point for hybrid pipelines where our pseudo-labels can be used to bootstrap downstream supervised training. The experiments also highlight a critical limitation in applying general foundation models to specialized material CT data. Instead of merely missing features, the foundation model fails to maintain instance-level structural integrity. It simultaneously hallucinates instances from background noise and merges distinct, adjacent particles into artificially large structures. In contrast, our model-free framework captures a far more comprehensive representation of the material features.

The practical value of this structural integrity is evidenced by our downstream validation. In microstructural quantification, our model-free approach successfully extracted a filtered population of valid grains that closely tracked the physical sieve limits of the specimen across a $100-slice$ volumetric batch. The foundation model, on the other hand, struggles in the material characterization task. Furthermore, we demonstrated that a YOLOv11 model trained on pseudo-labels from our method achieved downstream segmentation performance remarkably comparable to a model trained on labels from SAM at a zero-shot setting. This finding proves that our framework provides a sufficient density of high-fidelity training signals to enable robust feature learning in modern detection architectures, making it a highly practical solution for the ICME workflow.

The transparency of our approach remains a primary advantage over DL alternatives. We can trace specific prediction artifacts back to tunable parameters in our algorithm. This provides a clear and interpretable pathway for domain-specific adaptation. This transparency stands in stark contrast to the black-box nature of foundation models, where diagnosing and mitigating errors is a far more complex task.

Furthermore, the computational efficiency of our method is significant. Our model-free algorithm generated labels in an average of $15$ seconds per image on a CPU. This is nearly twice as fast as the SAM approach, which required $28$ seconds despite the use of GPU acceleration. This dramatic reduction in processing time and hardware dependency makes advanced image analysis more accessible to environments that lack high-performance computing resources.

Our framework provides practical value in two distinct ways. It serves as an invaluable tool for rapid prototyping and establishing performance baselines. Any subsequent deep learning model must justify its increased computational budget by demonstrating a significant performance gain over this efficiently produced benchmark. It also provides an excellent starting point for hybrid pipelines where our generated labels can be used to pre-train a model before fine-tuning on a small and expert-annotated dataset.

\section{Limitations and Future Works} \label{sec:pseudo-limitations-and-future-works}
While our model-free framework offers computational efficiency and interpretability, its limitations suggest several avenues for future research.

A primary constraint is the framework's sensitivity to image quality. Reliance on pixel intensity renders the process vulnerable to CT artifacts, such as beam hardening and electronic noise. Although our composite approach mitigates some sensitivity, datasets with significant artifacts remain challenging for accurate segmentation. Furthermore, the algorithm is optimized for discrete objects and may struggle with complex morphologies, including interwoven fibers or interconnected pores. Such structures require specialized modifications to the region-growing logic to ensure precise boundary identification.

The sensitivity analysis presented in Section~\ref{sec:pseudo-sensitivity-analysis} provides a concrete characterization of this tuning requirement. On the Dendrites dataset, the intensity similarity threshold $\tau$ exhibits a stable operational plateau spanning at least four units below the baseline value, with performance degrading rapidly for values above the baseline. This plateau behavior is specific to datasets where the target phase exhibits relatively uniform internal intensity, such as the solid dendritic structures examined here. For materials with lower phase contrast, heterogeneous internal intensity profiles, or significant beam hardening artifacts, the stability plateau may be narrower, requiring more careful threshold selection and potentially limiting the applicability of a single fixed threshold across a full dataset. Importantly, this narrowing of the stability plateau does not translate into unpredictable behavior. As established in Section \ref{sec:pseudo-sensitivity-analysis}, the transition point remains tied to the underlying phase contrast in the image, so the appropriate operating range can still be identified through direct histogram inspection even for datasets with a narrower margin for error. Similarly, the dilation kernel size, which governs the gap-closing behavior in Section \ref{sec:pseudo-label-generation}, may require adjustment for materials with finer microstructural features or thinner boundary regions than those examined in this work.

Downstream quantitative detection validation on the Dendrites dataset (\ref{sec:pseudo-validation-dataset}) is further limited by its test split, which as released by the dataset source contains only two images, too small a sample to support a statistically meaningful comparison of YOLO models trained on our pseudo-labels against those trained on SAM-generated labels. Expanding this test split, or supplementing it with additional annotated slices, would enable a more robust downstream detection comparison in future work.

\section{Conclusion} \label{sec:pseudo-conclusion}
In this work, we addressed the challenge of generating high-quality pseudo-labels for material CT segmentation in data-scarce environments. We introduced a multi-stage model-free framework that combines classical image processing techniques to produce reliable annotations without the computational overhead or opaque nature of DL methods. Quantitative evaluation demonstrated that our algorithm achieves substantially higher overlap and recall than the zero-shot foundation model, and stronger instance-level structural integrity than either the foundation model or the traditional Watershed algorithm at matched noise-removal thresholds.

Qualitative experiments further demonstrated that models trained with our pseudo-labels achieve object detection and instance segmentation performance comparable to those trained using Segment Anything Model (SAM) annotations. A primary advantage of our approach is its computational efficiency. Our framework generates pseudo-labels in nearly half the time per image on a standard CPU, whereas zero-shot SAM-based methods require twice the duration despite utilizing GPU acceleration. This reduction in processing time and hardware requirements provides a fast, affordable solution to the pseudo-labeling bottleneck, particularly for environments lacking high-performance computing resources.

Our method serves as a pragmatic tool for rapid model prototyping and establishing performance baselines. The transparency of the algorithm allows for domain-specific adaptation and straightforward error analysis. By facilitating accurate whole-grain annotations, this framework enables more reliable microstructural quantification, specifically regarding grain count, morphology, and size distribution. These high-fidelity digital descriptors provide a critical link between rapid pseudo-labeling and the precise characterization required for advanced material informatics and predictive modeling.

\section*{Data Availability}
The primary dataset (Section \ref{sec:pseudo-primary-dataset}) is owned by \citet{Imseeh2020-az} and can be requested from the authors. The validation dataset (Section \ref{sec:pseudo-validation-dataset}) is owned by \citet{Stan2020-ys} and can be accessed from \citet{Stan2019-kg}. The source code used for the experiment is available on \href{https://github.com/DATALab-NIU/Model-free-pseudo-label-generation}{GitHub}.

\section*{Conflict of Interest}
The authors declare that they have no conflicts of interest.

\bibliography{references}

\end{document}